\documentclass[lettersize,journal]{IEEEtran}
\usepackage{amsmath,amsfonts}
\usepackage{algorithmic}
\usepackage{algorithm}
\usepackage{array}
\usepackage[caption=false,font=normalsize,labelfont=sf,textfont=sf]{subfig}
\usepackage{textcomp}
\usepackage{stfloats}

\usepackage{verbatim}
\usepackage{graphicx}
\usepackage{cite}
\usepackage[normalem]{ulem}
\usepackage{multirow}
\usepackage{hhline}
\usepackage{hyperref}
\hypersetup{hypertex=true,
 colorlinks=true,
 linkcolor=blue,
 anchorcolor=blue,
 citecolor=blue}
\usepackage{makecell}
\usepackage{bbding}
\usepackage{multirow,booktabs}
\usepackage{url}
\usepackage[T1]{fontenc}

\begin{document}

\title{Three-in-One: Robust Enhanced Universal Transferable Anti-Facial Retrieval in Online Social Networks}

\author{Yunna Lv*, Long Tang*, Dengpan Ye\dag,~\IEEEmembership{Member,~IEEE}, Caiyun Xie, Jiacheng Deng and Yiheng He

\thanks{Yunna Lv, Long Tang, Dengpan Ye, Caiyun Xie, Jiacheng Deng and Yiheng He are with Wuhan University, School of Cyber Science and Engineering, Key Laboratory of Aerospace Information Security and Trusted Computing, Ministry of Education, Wuhan, 430072, China. (e-mails: \{lvyunna, l\_tang, yedp, caiyunxie, dengjiacheng, yihenghe\}@whu.edu.cn)}
\thanks{* Yunna Lv and Long Tang contribute equally to this paper.}
\thanks{† Corresponding author.}
}



\maketitle

\begin{abstract}
Deep hash-based retrieval techniques are widely used in facial retrieval systems to improve the efficiency of facial matching. However, it also carries the danger of exposing private information. Deep hash models are easily influenced by adversarial examples, which can be leveraged to protect private images from malicious retrieval. The existing adversarial example methods against deep hash models focus on universality and transferability, lacking the research on its robustness in online social networks (OSNs), which leads to their failure in anti-retrieval after post-processing. Therefore, we provide the first in-depth discussion on robustness adversarial perturbation in universal transferable anti-facial retrieval and propose Three-in-One Adversarial Perturbation (TOAP). Specifically, we construct a local and global Compression Generator (CG) to simulate complex post-processing scenarios, which can be used to mitigate perturbation. Then, we propose robust optimization objectives based on the discovery of the variation patterns of model's distribution after post-processing, and generate adversarial examples using these objectives and meta-learning. Finally, we iteratively optimize perturbation by alternately generating adversarial examples and fine-tuning the CG, balancing the performance of perturbation while enhancing CG's ability to mitigate them. Numerous experiments demonstrate that, in addition to its advantages in universality and transferability, TOAP significantly outperforms current state-of-the-art methods in multiple robustness metrics. It further improves universality and transferability by 5\% to 28\%, and achieves up to about 33\% significant improvement in several simulated post-processing scenarios as well as mainstream OSNs, demonstrating that TOAP can effectively protect private images from malicious retrieval in real-world scenarios.
\end{abstract}


\section{Introduction}
\begin{figure}[tb]
 \centering
 \includegraphics[width=1\linewidth]{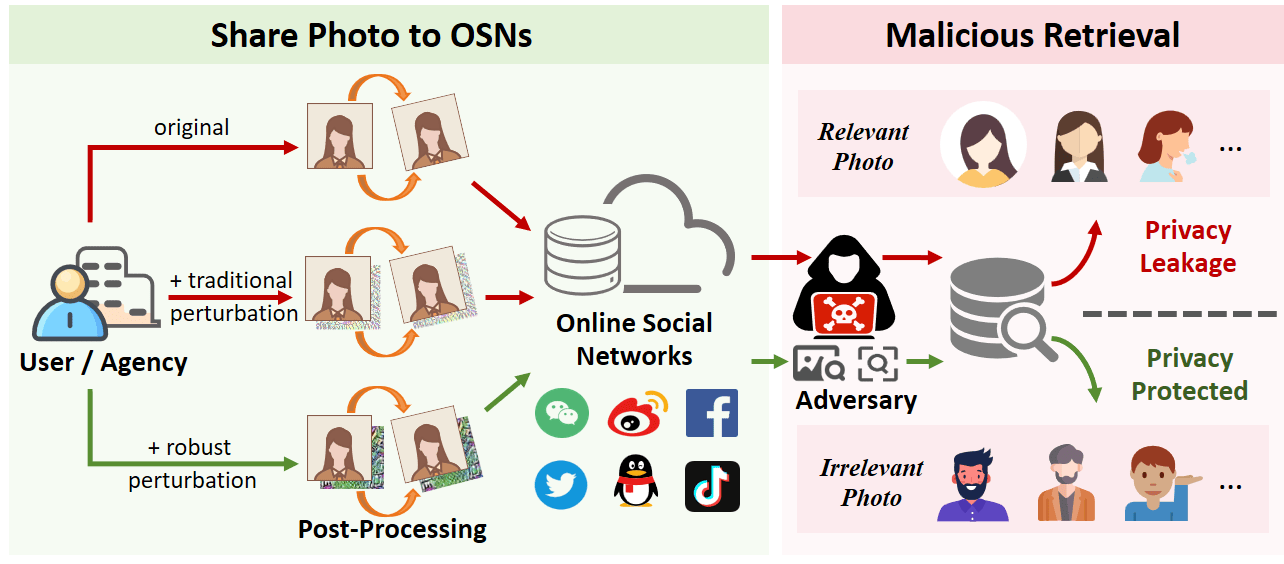}
 \caption{Threat Scenario. The red arrows indicate that adversary utilizes a photo containing user’s identity to retrieve user’s relevant photos, which leads to privacy leakage, such as user's daily life, job, and health condition. The green arrows indicate that adversary is unable to retrieve user’s relevant photos using user’s photo with robust perturbation, thus protecting user’s privacy.}
 \label{fig1}
\end{figure}

\IEEEPARstart{W}{ith} the advancement of the Internet, artificial intelligence and big data have become two main pillars driving social change. In order to extract desired information from big data, some companies collect facial images to train facial retrieval systems. Deep hash techniques, which combine deep expression and efficient retrieval capabilities in image retrieval tasks, are widely used in facial retrieval systems. However, while deep hash-based retrieval techniques bring convenience, they also have security risks. As shown by the top red arrows in Fig. \ref{fig1}, user or user information collection agency publicly posts photo containing user's information to online social networks (OSNs). OSNs will first perform post-processing on the photo to reduce the amount of data transfer and speed up loading \cite{wang2020towards,qu2024df}. Then, illegal user (adversary) can maliciously retrieve relevant photos of the user through this photo, leading to privacy leakage and potentially facilitating criminal activities such as telecommunication fraud and cyber violence. Therefore, there is a need to focus on privacy protection in deep hash-based retrieval systems.

Deep hash models inherit vulnerabilities to adversarial attacks \cite{xiao2020evade,zhao2023precise,hu2021advhash,tang2024once}. Therefore, adversarial examples can be used for facial anti-retrieval. As shown by the green arrows in Fig. \ref{fig1}, the user or the user information collection agency adds an adversarial perturbation to the photo before uploading it to OSNs, preventing the adversary from retrieving other private images, even after post-processing. However, adversarial attacks in real-world face three challenges, including cross-image universality, cross-model transferability, and cross-image post-processing robustness. Firstly, generating specific perturbations for different images is time-consuming, which makes the use of universal adversarial perturbations more essential in real-world scenarios. Secondly, companies usually do not disclose the architectures of deep hash model for commercial confidentiality and security reasons, making the transferability of adversarial perturbation crucial. Most importantly, due to the application of post-processing in OSNs, such as JPEG \cite{shin2017jpeg}, existing adversarial example methods may be invalid for adversarial features eliminated by these operations \cite{wang2020towards,qu2024df}. This means that adversary can still utilize these post-processed adversarial examples to retrieve relevant user images, thus making the user’s private images leaked, as shown by the middle red arrows in Fig. \ref{fig1}. Therefore, we must take into account the robustness against image post-processing to ensure that adversarial examples can work well in real-world.

\begin{figure}[tb]
 \centering
 \includegraphics[width=0.98\linewidth]{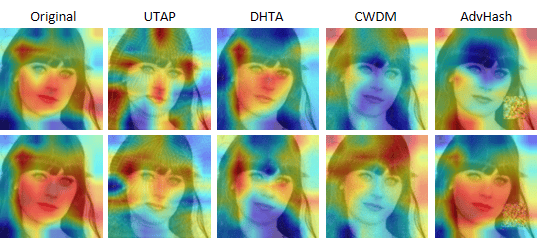}
 \caption{Visualization of the change in model's focus areas before (line 1) and after (line 2) JPEG compression (quality factor is 60). The experimental setting is CASIA, DHD, VGG16.}
 \label{fig2}
\end{figure}

Recently, researchers have extensively studied the universality and transferability of unimodal adversarial examples \cite{xiao2020evade,zhao2023precise,hu2021advhash,tang2024once} as well as cross-modal adversarial examples \cite{zhang2023proactive
,zhu2023efficient,wang2023targeted}. Among them, \cite{xiao2020evade} and \cite{tang2024once} have conducted a series of studies on privacy preservation in deep hash-based retrieval systems. However, there is still no method that has thoroughly discussed the robustness of adversarial examples against deep hash models in OSNs. Moreover, the effectiveness of existing methods is mitigated after image post-processing operations because of the significant change in the model's focus areas, as shown in Fig. \ref{fig2}. Therefore, there is an urgent need for a more robust adversarial perturbation in universal transferable anti-facial retrieval to enhance the practicality in real-world.

To overcome practical challenges, we propose \textbf{T}hree-in-\textbf{O}ne \textbf{A}dversarial \textbf{P}erturbation (TOAP), which is the first OSNs-oriented robust, universal, and transferable adversarial example method against deep hash-based retrieval systems. Specifically, we introduce a local and global Compression Generator (CG), inspired by JPEG compression \cite{shin2017jpeg}, which first splits the adversarial examples into grids for local post-processing, and then merges the grids for global post-processing, effectively simulating the image post-processing operations in OSNs. Then, we find robust optimization objectives that are almost unrelated to image post-processing for optimizing the generation of adversarial examples, including original cluster centers and data space centers. Finally, we alternately generate adversarial perturbation using original as well as local and global CG-processed perturbed images, and fine-tune the CG using feature-level, pixel-level, and hash-level optimization losses to minimize the impact of perturbation. It effectively balances the robustness as well as universality and transferability of adversarial examples and enhances the CG's ability to simulate image post-processing operations simultaneously.

The main contributions are summarized as follows:
\begin{itemize}
\item We first propose OSNs-oriented robust adversarial perturbation TOAP in universal transferable anti-facial retrieval. Excellent robustness ensures the adversariality of adversarial examples when post-processed by OSNs, combining with universality and transferability, effectively protecting users' privacy.
\item We propose a local and global CG to process the adversarial examples, which effectively simulates the image post-processing operations in OSNs and facilitates the generation of more robust adversarial perturbation.
\item We propose an attack strategy from point to space, taking cluster centers and data space as optimization objectives, respectively. By alternately optimizing adversarial examples using original as well as local and global CG-processed perturbed images, and fine-tuning CG using pixel-level, feature-level, and hash-level losses, we enhance the performance of perturbation while strengthening the CG's ability to simulate post-processing of OSNs.
\item We perform experiments using different algorithms, models, and image post-processing operations on facial datasets CASIA \cite{yi2014learning} and VGGFace2\cite{cao2018vggface2}, to comprehensively validate TOAP’s robustness as well as universality and transferability. Experimental results show that TOAP further improves universality and transferability by 5\% to 28 \%, robustness by up to 33\% after simulated post-processing and up to 10\% in mainstream OSNs, including Facebook, WeChat, and Weibo. Moreover, TOAP also demonstrates effectiveness on real-world platforms, such as Google Images and 360 Images.
\end{itemize}

\section{RELATED WORKS}
\subsection{Deep Hash-Based Image Retrieval}
Deep hash algorithms have been applied to image retrieval to achieve efficient retrieval of large-scale image data. It typically involves training a Convolutional Neural Networks (CNN), such as VGG \cite{simonyan2014very} or ResNet \cite{he2016deep}, and a fully connected layer to transform the original high-dimensional feature space into a binary Hamming space. The similarity between query and database images is then measured using their Hamming distance. HashNet \cite{cao2017hashnet} is a classical deep hash algorithm, which utilizes a continuation method with convergence, ensuring accurate learning of binary hash codes. However, due to the less concentrated data distribution in its training, the accuracy is limited. To address this problem, some deep hash algorithms guided by data centers are proposed. CSQ \cite{yuan2020central} is a highly clustered deep hash algorithm that encourages hash codes of similar data to be close to a common center, thereby improving the accuracy of retrieval. DHD \cite{jang2022deep} is a robust high-precision deep hash algorithm that uses self-distillation hash algorithm to learn transformed images and employs hash proxy-based similarity learning to train hash codes. We evaluate TOAP based on the highly clustered and high-precision algorithms, i.e., CSQ and DHD.

\subsection{Adversarial Attacks}
Adversarial examples are created by adding well-designed perturbations to original data, which can cause Deep Neural Networks (DNNs) to make incorrect predictions. It has been widely applied to different models \cite{goodfellow2014explaining,wei2022towards,xu2024perturbing}, different tasks \cite{qu2024df,qian2024enhancing,sun2024task,aafaq2022language}, different modalities \cite{mkadry2017towards,li2023voice,wei2022cross}. Early methods usually update adversarial examples by gradient obtained from backpropagation \cite{goodfellow2014explaining,kurakin2018adversarial,mkadry2017towards}. Subsequently, some researchers focus on improving the transferability of adversarial examples and propose a series of effective methods \cite{dong2018boosting,lin2019nesterov,fang2022learning}. Then, universal adversarial perturbations (UAP) \cite{moosavi2017universal} has been proposed to enhance the generalization of adversarial examples, which can successfully attack DNNs by adding the same perturbation to different images. Recently, researchers simulate image post-processing operations to model OSNs, improving the robustness of adversarial examples. For example, Qu et al. \cite{qu2024df} combined U-Net \cite{isola2017image} with JPEG compression layers \cite{shin2017jpeg}, training an encoding-decoding structure to simulate various post-processing operations performed by OSNs.

\subsection{Adversarial Attacks against Image Retrieval}
DNNs-based image retrieval is also susceptible to adversarial attacks. For attacks on deep feature-based retrieval models, Liu et al. \cite{liu2019s} propose an untargeted adversarial example method. Subsequently, UAA-GAN \cite{zhao2019unsupervised} and AP-GAN \cite{zhao2022ap} are proposed, which are GAN-based adversarial perturbation and patch method, respectively. Tolias et al. \cite{tolias2019targeted} propose TMAA, which is the first targeted adversarial example for image retrieval. Chen et al. \cite{chen2021dair} propose DAIR, a query-based black-box attack method. Li et al. \cite{li2019universal} propose UAP to enhance the generalizability of the adversarial examples. 

For attacks on deep hash-based retrieval models, Yang et al. \cite{yang2018adversarial} propose adversarial examples for the first time. Xiao et al. \cite{xiao2020evade} propose CWDM, an untargeted adversarial examples method, aiming to protect image privacy. Bai et al. \cite{bai2020targeted} propose DHTA, which is the first targeted adversarial attack. Lu et al. \cite{lu2021smart} propose a smart deep hash attack method, which effectively improves the image quality of adversarial examples. Zhao et al. \cite{zhao2023precise} propose a precise target-oriented attack to improve the performance of targeted adversarial attacks. Xiao et al. \cite{xiao2021you} propose a targeted attack method to further explore the transferability of adversarial examples. Then, Hu et al. \cite{hu2021advhash} and Tang et al. \cite{tang2024once} propose a class-wise universal adversarial patch method and a universal adversarial perturbation method, respectively, improving the generalizability of adversarial examples. Recently, adversarial attacks based on cross-modal hash retrieval \cite{zhang2023proactive,zhu2023efficient,wang2023targeted} have also gained widespread attention. However, since the scenarios of these methods are different from this paper, no comparison will be made.

Although existing methods have made some progress in specific scenarios, they still face significant challenges in achieving the balance of robustness, universality, and transferability of adversarial examples, resulting in the inability to work well in real-world. Therefore, TOAP will focus on the balance of the three properties. The existing adversarial attacks targeting image retrieval systems are shown in Table \ref{table1}.

\begin{table}
\centering
\setlength{\tabcolsep}{3.6pt}
\renewcommand{\arraystretch}{1.15}

\caption{The comparison of existing adversarial attacks against image retrieval.}
\label{table1}
\begin{tabular}{cccccc} 
\Xhline{1.1pt}
\textbf{Method} & \textbf{Model} & \textbf{Form}& \textbf{Universal} & \textbf{Transferable} & \textbf{Robust} \\ 
\Xhline{1pt}
PIRE\cite{liu2019s} & Feature & Perturb & \XSolidBrush & \XSolidBrush & \textbf{\textit{Low}} \\
UAA-GAN\cite{zhao2019unsupervised} & Feature & Perturb & \XSolidBrush & \CheckmarkBold & \XSolidBrush \\
AP-GAN\cite{zhao2022ap} & Feature & Patch & \XSolidBrush & \XSolidBrush & \textbf{\textit{Low}} \\
TMAA\cite{tolias2019targeted} & Feature & Perturb & \XSolidBrush & \CheckmarkBold & \XSolidBrush \\
DAIR\cite{chen2021dair} & Feature & Perturb & \XSolidBrush & \XSolidBrush & \XSolidBrush \\
UAP\cite{li2019universal} & Feature & Perturb & \CheckmarkBold & \XSolidBrush & \XSolidBrush \\ 
\hline
HAG\cite{yang2018adversarial} & Hash & Perturb & \XSolidBrush & \CheckmarkBold & \XSolidBrush \\
CWDM\cite{xiao2020evade} & Hash & Perturb & \XSolidBrush & \CheckmarkBold & \XSolidBrush \\
DHTA\cite{bai2020targeted} & Hash & Perturb & \XSolidBrush & \XSolidBrush & \XSolidBrush \\
SDHA\cite{lu2021smart} & Hash & Perturb & \XSolidBrush & \XSolidBrush & \XSolidBrush \\
PTA\cite{zhao2023precise} & Hash & Perturb & \XSolidBrush & \CheckmarkBold & \XSolidBrush \\
NAG\cite{xiao2021you} & Hash & Perturb & \XSolidBrush & \CheckmarkBold & \XSolidBrush \\
AdvHash\cite{hu2021advhash} & Hash & Patch & \textit{\textbf{Class-wise}} & \XSolidBrush & \XSolidBrush \\
UTAP\cite{tang2024once}& Hash & Perturb & \CheckmarkBold & \CheckmarkBold & \textbf{\textit{Low}} \\
\textbf{TOAP} & \textbf{Hash} & \textbf{Perturb} & \textbf{\CheckmarkBold} & \textbf{\CheckmarkBold} & \textbf{\textit{High}} \\
\Xhline{1.1pt}
\end{tabular}
\end{table}

\section{METHODOLOGY}
\subsection{Overview}
\begin{figure}[tb]
 \centering
 \includegraphics[width=1.0\linewidth]{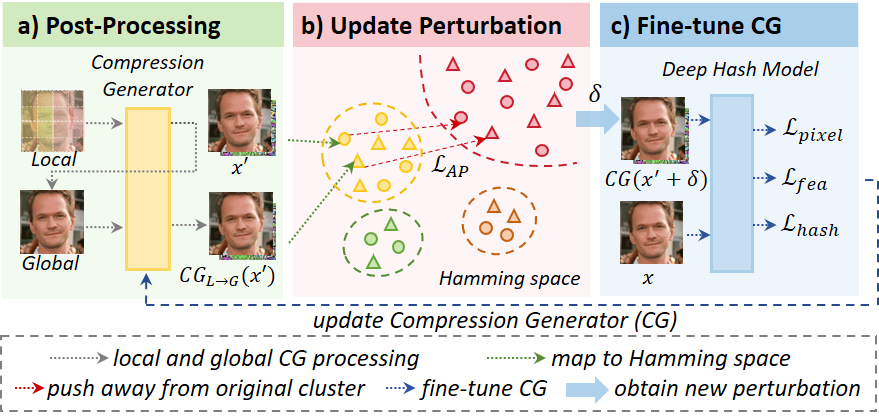}
 \caption{The pipeline of TOAP. a) using local and global CG to process adversarial examples; b) optimizing perturbation from point to space using meta-learning; c) fine-tuning CG using pixel-, feature, and hash-level losses.}
 \label{pipeline}
\end{figure}

To improve the robustness of adversarial perturbation, we propose a local and global CG as shown in Fig. \ref{fig3}, simulating the image post-processing operations of OSNs. Then, we decompose the optimization task from point to space, defined as away from the original clusters and the data space, respectively, and the optimization objective is shown in Fig. \ref{fig6}. The adversarial perturbation is optimized by using the original as well as local and global CG-processed perturbed images to balance the adversariality and robustness of adversarial examples. Finally, following the training paradigms of \cite{huang2021initiative,van2023anti}, we iteratively optimize robust perturbation by alternately training adversarial examples and CG as shown in Fig. \ref{fig8}. This process balances the performance of adversarial examples while enhancing the CG’s ability to mitigate adversarial examples and simulate the image post-processing operations in OSNs. Fig. \ref{pipeline} displays the pipeline of TOAP.

\subsection{Preliminaries}
\subsubsection{Dataset}We define that $X={(x_i,y_i )}^N$ denotes that the dataset consists of $N$ images from $P$ users. $x_i$ denotes an image, $ y_i=(y_{i_1},...,y_{i_P})$ denotes the label of $x_i$, $y_{i_p} \in [0,1]$. If $ y_{i_p}=1 $, then the image belongs to user $p$. Suppose that the user $p$ has $N_p $ photos. Let $X_{tr} = (x_i, y_i)^{M_p \times P} $ denotes the training set with $M_p\times P$ images from $P$ users, where $M_p $ is far smaller than $N_p$.

\subsubsection{Deep Hash Model}We obtain the $K$-bit hash code of the image $x$ by the deep hash model $F(\cdot)$ as follows:
\begin{equation}
c = F(x) = sign(H(f(x))), \label{eq1}
\end{equation}
where $F(\cdot)$ consists of a feature layer $f(\cdot)$ and a $K$-dimensional fully connected layer $H(\cdot)$. The feature layer usually consists of a CNN model, e.g., VGG \cite{simonyan2014very} or ResNet \cite{he2016deep}. To optimize the computation of backpropagation, the model is usually trained using hyperbolic tangent function $tanh(\cdot)$ approximation to sign function $sign(\cdot)$, and the returned hash code is a $K$-bit continuous value ranging from -1 to 1.

\subsubsection{Similarity Metric}The hash codes $c_i$ of all images $x_i$ in database $X$ are calculated by Eq. \ref{eq1}. Given the query image $x_q$, its hash code $c_q$ can also be calculated by Eq. \ref{eq1}. And then, we can compute the Hamming distance between the query image and all database images using $c_q$ and $c_i$:
\begin{equation}
D = \left\{ d_H(c_q, c_i) \right\}_{i=1}^{N} = \left\{ \frac{K - c_q \cdot c_i}{2} \right\}_{i=1}^{N}. \label{eq2}
\end{equation}
As a result, the deep hash model will ultimately return the ranking list of images in the database, ordered in ascending sequence based on their Hamming distances.

\subsubsection{Hash Centers}Existing methods \cite{hu2021advhash,tang2024once,bai2020targeted} typically generate adversarial examples by utilizing different hash centers as guidance. The hash centers can be obtained by voting using the hash codes of the training set images, including cluster centers $\{h_p\}_{i=1}^{P} = \{sign(\sum_{j=1}^{M_p} c_j )\}_{i=1}^{P}$, sub centers (data space centers) $\{h_s\}_{i=1}^{N_s} = \{sign(\sum_{p=r_{i_1}}^{r_{i_n}} h_p)\}_{i=1}^{N_s}$ and overall center $h_o = sign(\sum_{p=1}^{P} h_p )$, where $N_s$ is the number of sub centers selected randomly, and $\{r_{i_1},...,r_{i_n}\}$ denotes $n$ clusters selected randomly. 
\subsection{Local and Global Compression Generator}\label{sec1}
\begin{figure}[tb]
 \centering
 \includegraphics[width=1.0\linewidth]{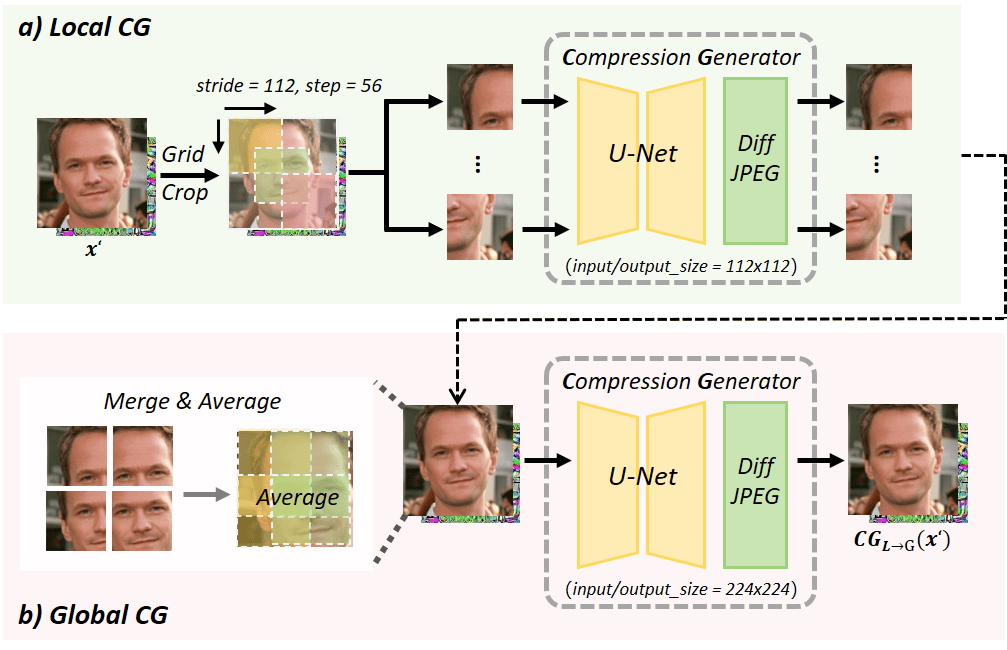}
 \caption{The pipeline of local and global CG.}
 \label{fig3}
\end{figure}

The most direct method for enhancing the robustness of adversarial examples is to upload them to OSNs, download the post-processed versions, and use them for training. However, uploading and downloading images is a time-consuming task, and excessive uploading and downloading operations may be recognized as abnormal behavior. Therefore, it is necessary to design a compression generator to simulate the image post-processing operations performed by OSNs. 

Given an adversarial example, the compression generator can output a post-processed adversarial example. Therefore, the compression generator is usually designed as an encoding-decoding image mapping structure. Specifically, we follow the settings in \cite{qu2024df,wu2022robust}, employing U-Net structure to learn the mapping relations of the images and using residual blocks to improve the efficiency of feature extraction. After that, we integrate a DiffJPEG \cite{shin2017jpeg} module to further simulate the image compression, resulting in the CG. The JPEG compression technique \cite{shin2017jpeg} employs a sequence of local and then global transformations on images, which guides us to simulate similar operations in OSNs by initially conducting localized image post-processing, followed by comprehensive global image post-processing with CG.

\subsubsection{Local CG}We divide the adversarial examples into multiple grids, treating each grid as a distinct sub-image for local image post-processing. These sub-images are then processed individually by CG, which is specifically designed to apply post-processing operations aimed at locally mitigating adversarial perturbations. By processing each grid independently, we are able to preserve localized details while minimizing the impact of adversarial perturbations in each section of the image. To ensure a smooth transition at the image boundaries, we use a sliding window with a predefined stride and step size to split the image into overlapping grids, which can help maintain continuity between adjacent grids.

\subsubsection{Global CG}We firstly merge the grids that have been post-processed locally for global image post-processing. Specifically, we calculate the average of the overlapping areas between adjacent grids to ensure the continuity and naturalness of the image. The merged images are subsequently input into CG for global post-processing, further simulating the image post-processing operations performed by OSNs and reducing the impact of adversarial perturbation through more comprehensive image post-processing. Specific implementation details are shown in Fig. \ref{fig3}.

\subsection{Optimization from Point to Space}\label{sec2}

\begin{figure}[tb]
\centering
\subfloat{\includegraphics[width=1.71in]{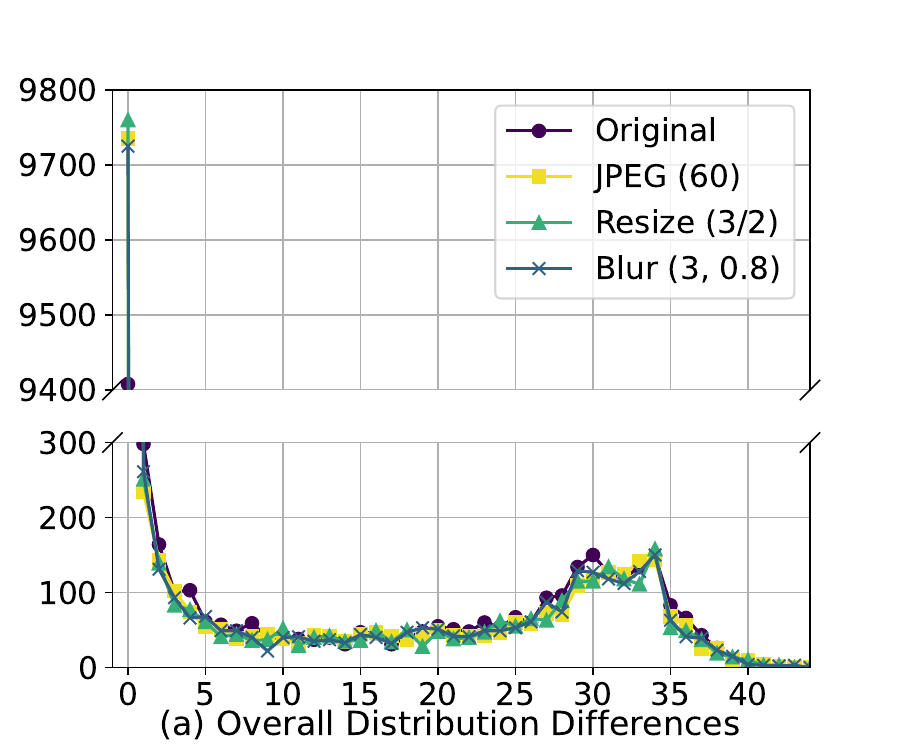}}\hspace{0in}
\subfloat{\includegraphics[width=1.71in]{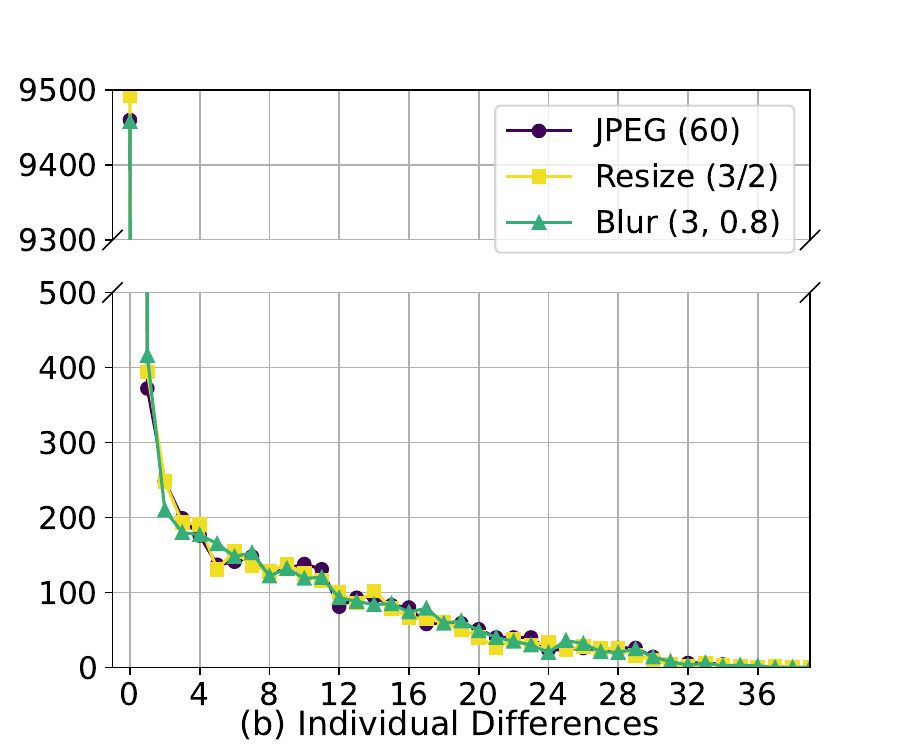}}
\caption{The relationship between the number of database samples and the Hamming distance between: (a) samples and their cluster centers; (b) original samples and post-processed samples. Y-axis denotes the number of samples and X-axis denotes the Hamming distance. The experimental setting is CASIA, DHD, ResNet50. The quality factor of JPEG compression is 60, the resize ratio is 3/2, and the Gaussian kernel size for Gaussian blur is 3 with a standard deviation of 0.8.}
\label{fig4}
\end{figure}

\begin{figure}[tb]
\centering
\subfloat{\includegraphics[width=1.71in]{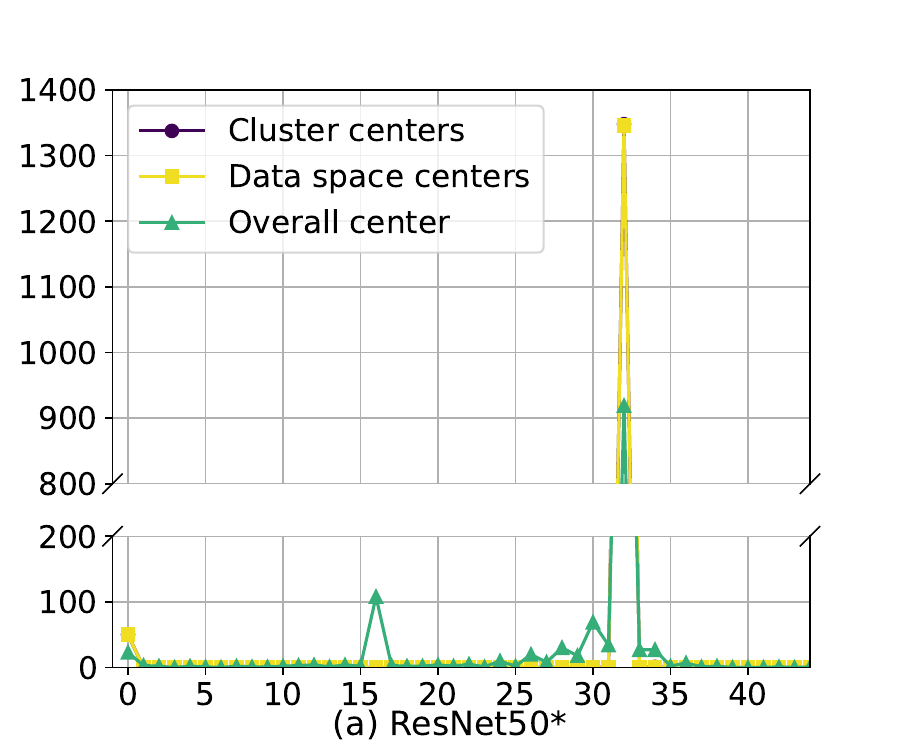}}\hspace{0in}
\subfloat{\includegraphics[width=1.71in]{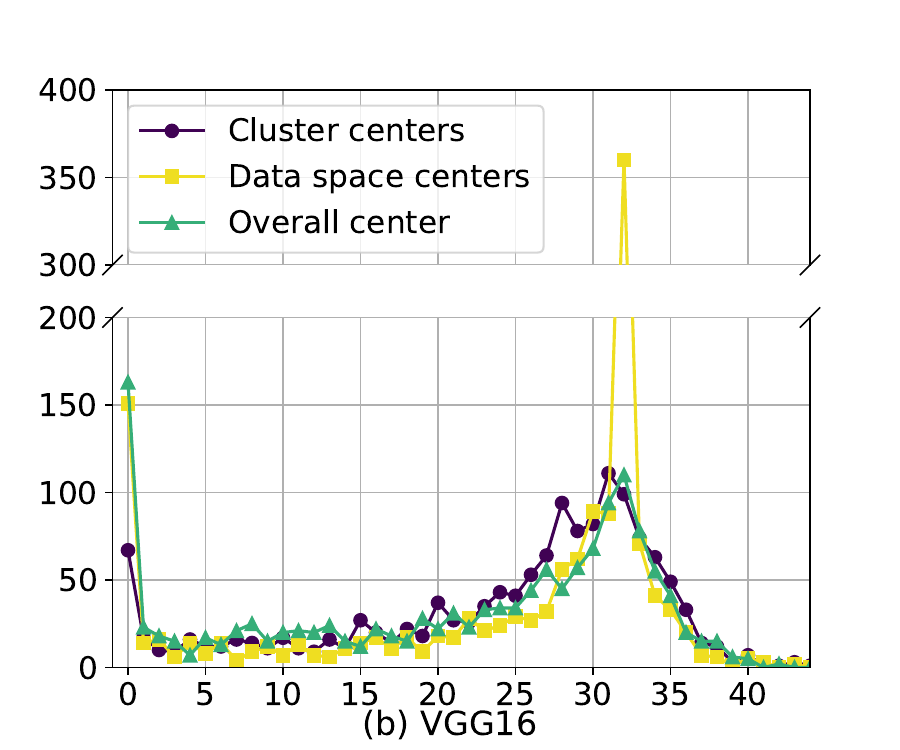}}
\caption{The relationship between the number of adversarial examples guided by different centers and the Hamming distance to their original cluster centers. A larger number of adversarial examples deviating from 0 indicates stronger adversariality. The experimental setting is CASIA, CSQ, ResNet50. * denotes the white-box adversarial results.}
\label{fig5}
\end{figure}

\begin{figure}[tb]
 \centering
 \includegraphics[width=0.95\linewidth]{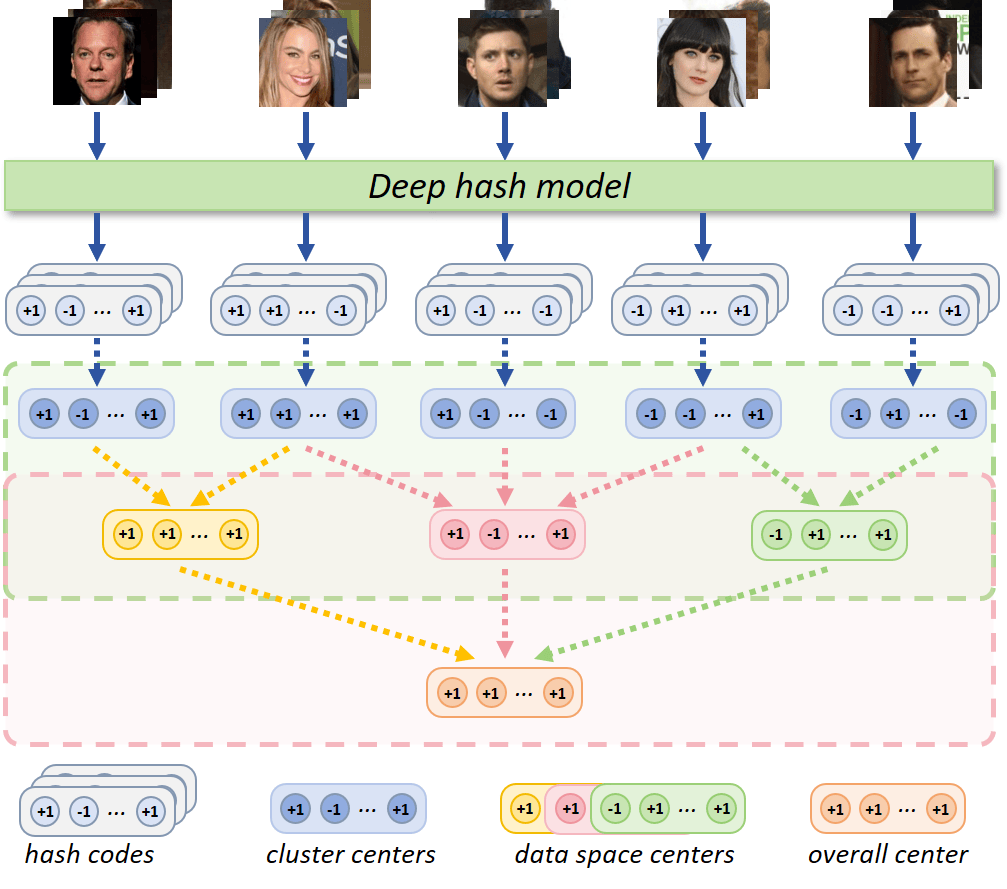}
 \caption{The illustration of the selection of different centers. The red shaded area represents the objective selection method of UTAP \cite{tang2024once}, while the green shaded area represents that of our TOAP. The dashed line indicates voting.}
 \label{fig6}
\end{figure}

Inspired by \cite{tang2024once}, we attempt to optimize adversarial examples by finding a fixed factor to enhance their robustness. We firstly observe that model's focus areas on original samples change little before and after post-processing, as shown in the first column of Fig. \ref{fig2}. Then, we visualize the relationship between the number of database samples and the Hamming distances of them to their cluster centers before and after image post-processing (as shown in Fig. \ref{fig4}(a)), as well as the Hamming distances between original samples and post-processed samples (as shown in Fig. \ref{fig4}(b)) to find suitable optimization objectives. The results indicate that the overall data distribution changes little before and after post-processing, and most of individual sample positions change minimally, suggesting high similarity between the data distributions before and after post-processing. Therefore, we optimize the adversarial perturbation using data centers related to the data distribution, which are almost unrelated to the post-processing operations, to improve the robustness of adversarial perturbation. 

Fig. \ref{fig5} illustrates the impact of different centers on the optimization of adversarial examples by visualizing the relationship between the number of adversarial examples guided by different centers and the Hamming distance to their original cluster centers. Fig. \ref{fig5}(a) shows the white-box results, and Fig. \ref{fig5}(b) shows the black-box results. It can be observed that, in the white-box case, using cluster centers $h_p$ and data space centers $h_s$ help the adversarial examples move further away from the original clusters, while using the single overall center $h_o$ leads to optimization difficulty. Similarly, in the black-box case, the adversarial examples optimized with $h_p$ exhibit less clustering around their original clusters, and those optimized with $h_s$ tend to cluster more distantly from the original clusters. In contrast, the adversarial examples guided by $h_o$ do not effectively deviate from their original clusters. 



Based on the above observations, we propose a novel objective combining cluster centers and data space centers, as shown in the green shaded area of Fig. \ref{fig6}, addressing the optimization challenge of using a single overall center as objective in \cite{tang2024once}, as shown in the red shaded area of Fig. \ref{fig6}. Then, we utilize meta-learning to optimize different objectives following \cite{tang2024once}, transforming the problem into maximizing the Hamming distance between adversarial examples and cluster centers (meta-training) as well as data space centers (meta-testing) for a more refined optimization process. The optimization method is defined as follows:
\begin{align}
&\max_{\delta} D(F(X_{tr}'), h) \Rightarrow \max_{\delta} \sum_{j=1}^{M_p \times P} d_H(sign(H(x_j')), h), \notag \\
& \quad h = 
\begin{cases} 
h_p, & \text{in meta-training,} \\
h_s, & \text{in meta-testing,}
\end{cases}
\text{ s.t.} \quad \|x' - x\|_\infty \leq \epsilon, \label{eq3}
\end{align}
where $\delta$ denotes the universal adversarial perturbation, $x_j$ denotes the images from $X_{tr}$, and $X_{tr}' $ consists of adversarial examples $x_j'=x_j +\delta$. We use the $l_\infty $ norm to constrain the magnitude of perturbation.

Then, we optimize the adversarial perturbation using both the original perturbed images $x_j'$ as well as the local and global CG-processed perturbed images $CG_{L \rightarrow G}(x_j')$, balancing the adversariality and robustness of the adversarial examples. Additionally, the $sign(\cdot)$ function is replaced with $tanh(\cdot)$ to compute continuous gradients. Finally, the loss function is defined as follows:
\begin{align}
\mathcal L_{AP}(X&_{tr}, \delta, h, CG)= \alpha\cdot
\frac{\sum_{j=1}^{M_p \times P} (K-h^T \cdot tanh(H(x_j')))}{2\times M_p \times P}\notag\\
&+\beta \cdot \frac{\sum_{j=1}^{M_p \times P} (K-h^T \cdot tanh(H(CG_{L\rightarrow G}(x_j'))))}{2\times M_p \times P},\notag\\
& \quad h = 
\begin{cases} 
h_p, & \text{in meta-training,} \\
h_s, & \text{in meta-testing,}
\end{cases}
\label{eq4}
\end{align}
where $\alpha$ and $\beta$ are hyperparameters. Then, we compute the meta grads at the $t$-th iteration ($t\in T$) as follows:
\begin{align}
grad_i &= \nabla_{\delta} \mathcal L_{AP} (X_{tr}, \delta, h, CG),\notag\\
\quad i, \delta, h &= 
\begin{cases} 
1, {{\delta}^t}, h_p, & \text{in meta-training,} \\
2, {{\delta}^t}', h_s, & \text{in meta-testing,}
\end{cases}
\label{eq5}
\end{align}
\begin{equation}
{{\delta}^t}' = clip_{\epsilon} ({\delta}^t + \epsilon \cdot sign(grad_1)).
\label{eq6}
\end{equation}

Finally, we update the adversarial perturbation based on meta grads $grad_1$ and $grad_2$ with the learning rate $\eta$:
\begin{equation}
\delta^{t+1} = clip_\epsilon (\delta^t + \eta \cdot sign(grad_1 + grad_2)). \label{eq7}
\end{equation}

\begin{figure}[tb]
 \centering
 \includegraphics[width=1.0\linewidth]{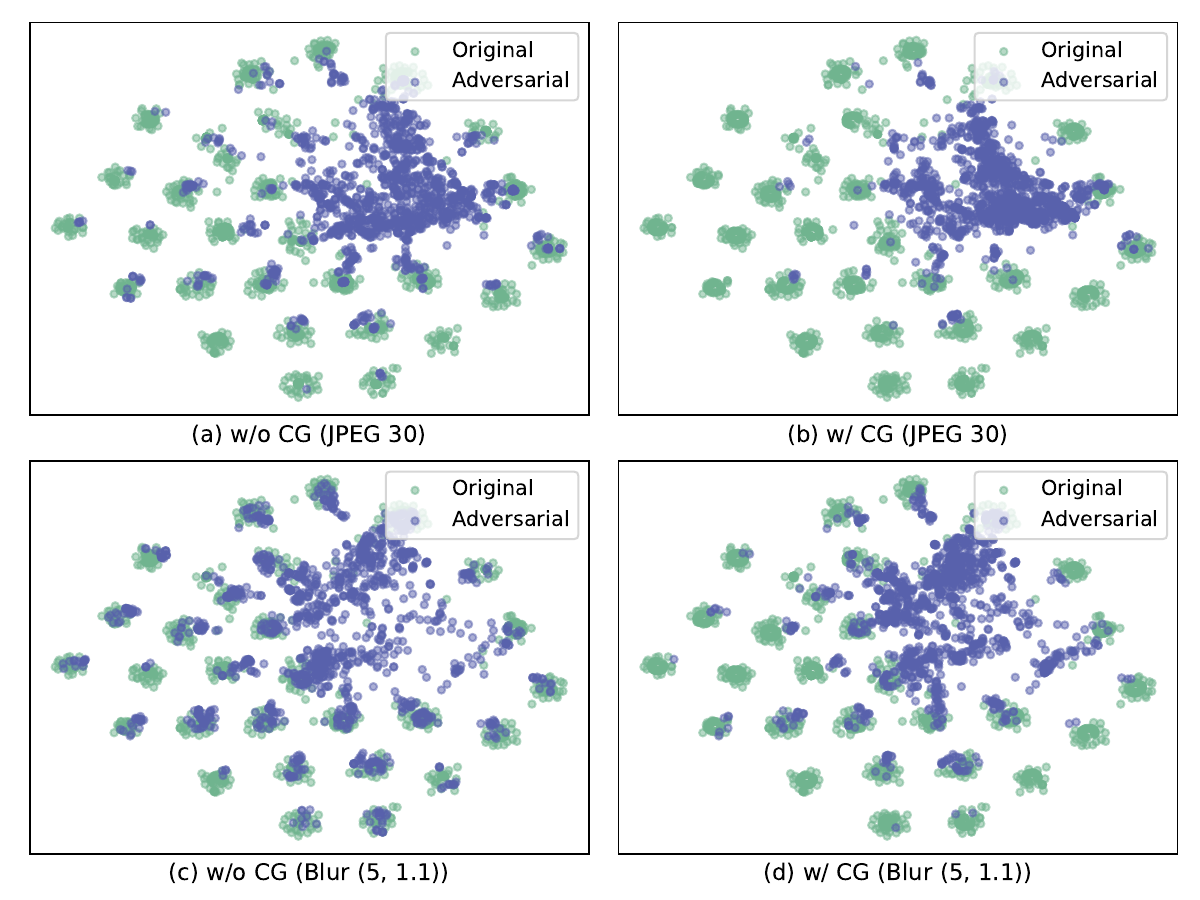}
 \caption{T-SNE distributions of original/adversarial examples before and after JPEG compression or Gaussian blur. W/o CG means optimizing adversarial examples using only the original image, while w/ CG means optimizing them by introducing local and global CG. The experimental setting is CASIA, DHD, ResNet50. The quality factor of JPEG compression is 30 and the Gaussian kernel size for Gaussian blur is 5 with a standard deviation of 1.1.}
 \label{fig7}
\end{figure}

We also demonstrate the improvement in the robustness of adversarial examples due to the introduction of local and global CG through the t-SNE plots, as shown in Fig. \ref{fig7}. It displays the distributions of post-processed original images and adversarial examples trained for 20 epochs with and without local and global CG. It is observed that the adversarial examples trained with local and global CG are able to cluster more efficiently and stay farther away from the original clusters when facing different post-processing operations, such as JPEG compression and Gaussian blur.

\subsection{Fine-tune Grid Compression Generator}\label{sec3}
OSNs typically mitigate adversarial perturbations. Therefore, we construct the loss function to fine-tune the CG at pixel-level, feature-level, and hash-level, which further makes the adversarial examples as similar as possible to the original clean samples and effectively simulates the operations of OSNs. This process not only mitigates the impact of adversarial perturbation, but also promotes the generation of a more robust perturbation.

For pixel-level, we utilize Mean Squared Error (MSE) loss to measure the pixel-level differences between the CG-processed adversarial examples and the original images, which is the most basic method to measure image differences:
\begin{equation}
\mathcal L_{pixel}(X_{tr}, \delta, CG) = \frac{\sum_{j=1}^{M_p \times P} [CG(x_j') - x_j]^2}{M_p \times P}. \label{eq8}
\end{equation}

For feature-level, we also use the MSE loss to measure the discrepancies between the features of the CG-processed adversarial examples and the original images as follows:
\begin{equation}
\mathcal L_{fea}(X_{tr}, \delta, CG) = \frac{\sum_{j=1}^{M_p \times P} [f(CG(x_j')) - f(x_j)]^2}{M_p \times P}.\label{eq9}
\end{equation}
The feature-level loss focus on the higher-level feature representations of model, which can affect the model's decision-making. Minimizing the feature-level differences between the CG-processed adversarial examples and the original images can ensure their consistency in high-level features, which helps reduce the impact of perturbation on model’s decision-making.

For hash-level, we use the Hamming distance to measure the differences between the hash codes of the CG-processed adversarial examples and the original images:
\begin{align}
&\mathcal L_{hash}(X_{tr}, \delta, CG) = 
\notag \\ &\frac{ \sum_{j=1}^{M_p \times P} \left[ K - tanh(H(CG(x_j')))^T \cdot tanh(H(x_j)) \right]}{2 \times M_p \times P}. \label{eq10}
\end{align}
Deep hash models rely on the distance between images in the Hamming space for retrieval. Therefore, we minimize the Hamming distance between the CG-processed adversarial examples and the original images, ensuring that the adversarial examples stay within the original clusters and reducing the impact of adversarial perturbation.

\begin{figure}[tb]
 \centering
 \includegraphics[width=1\linewidth]{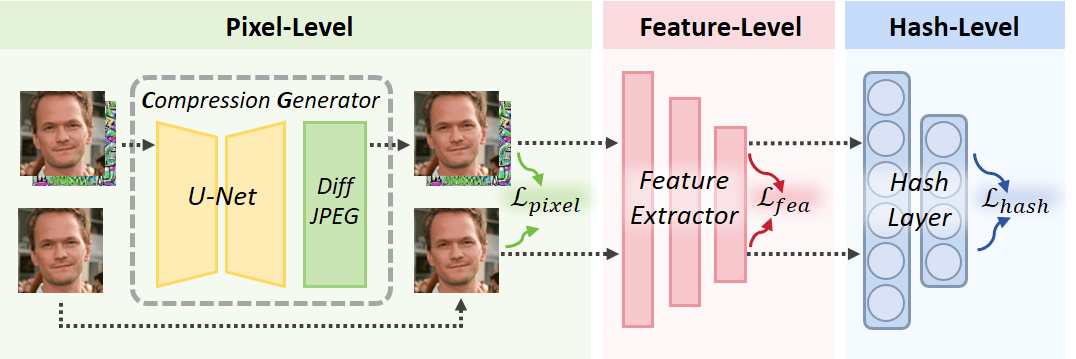}
 \caption{The illustration of fine-tuning CG. We fine-tune CG using the three-level losses, including pixel- (green), feature- (red), and hash-level loss (blue).}
 \label{fig8}
\end{figure}

Finally, the fine-tuning loss is determined as follows:
\begin{align}
\mathcal L&_{CG}(X_{tr}, \delta, CG) = \lambda_1 \cdot \mathcal L_{pixel}(X_{tr}, \delta, CG) \notag \\ & + \lambda_2 \cdot \mathcal L_{fea}(X_{tr}, \delta, CG) + \lambda_3 \cdot \mathcal L_{hash}(X_{tr}, \delta, CG), \label{eq11}
\end{align}
where $\lambda_1$, $\lambda_2$, and $\lambda_3$ are hyperparameters. By minimizing the loss function, we update the parameters of CG. The method for obtaining fine-tuning losses is illustrated in Fig. \ref{fig8}.

We optimize robust perturbation by alternately training the adversarial examples and the CG. The complete process is shown in Algorithm \ref{algorithm1}.

\begin{algorithm}[tb]
\caption{Three-in-One Adversarial Perturbation (TOAP)}
\label{algorithm1}
\textbf{Input}: training set $X_{tr}$; Compression Generator $CG(\cdot)$; cluster centers $\{h_p\}^{P}_{i=1}$; data space centers $\{h_s\}^{N_s}_{i=1}$; iteration number $T$; learning rate $\eta$.\\
\textbf{Output}: TOAP $\delta$.
\begin{algorithmic}[1] 
\STATE Initialize $\delta^0=0$, $mAP_{best}=1.0$, $CG_0=CG(\cdot)$.
\FOR{$t$ in $T$}
\STATE \textbf{// Generate Adversarial Perturbation}
\STATE Compute $\mathcal L_{AP}(X_{tr}, \delta ^t, h_p, CG_i)$ with Eq. \ref{eq4};
\STATE $grad_1 = \nabla_{\delta^t} \mathcal L_{AP} (X_{tr}, \delta^t, h_p, CG_i)$;
\STATE ${{\delta}^t}' = clip_{\epsilon} ({\delta}^t + \epsilon \cdot sign(grad_1))$;
\STATE Compute $\mathcal L_{AP}(X_{tr}, {\delta ^t}', h_s, CG_i)$ with Eq. \ref{eq4};
\STATE $grad_2 = \nabla_{{\delta^t}'} \mathcal L_{AP} (X_{tr}, {\delta^t}', h_s, CG_i)$;
\STATE $\delta^{t+1} = clip_\epsilon (\delta^t + \eta \cdot sign(grad_1 + grad_2))$;
\STATE Compute $mAP_i$ with Eq. \ref{eq12};
\STATE \textbf{// Fine-Tune CG}\;
\IF {$mAP_i \textless mAP_{best}$}
\STATE Compute $\mathcal L_{CG}(X_{tr},{\delta}^{t+1}, CG_i)$ with Eq. \ref{eq11};
\STATE Update $CG_{i+1};$
\STATE $mAP_{best}=mAP_i$;
\ELSE
\STATE $CG_{i+1}\leftarrow CG_{i}$;
\ENDIF
\ENDFOR
\STATE \textbf{return} $\delta^T$.
\end{algorithmic}
\end{algorithm}

\section{Experiments}
\subsection{Experimental Settings}
\subsubsection{Datasets}We conduct experiments using the facial datasets CASIA-WebFace (CASIA) \cite{yi2014learning} and VGGFace2 \cite{cao2018vggface2}. Since some of identities in the datasets contain a small number of images, we follow \cite{tang2024once} and select $P=28$ identities with $N_p>500$ in each dataset for the experiments to fulfill retrieval requirements. Specifically, we randomly select $100$ and $50$ images from each identity as training set and testing set, respectively, and the remaining images as database. Eventually, for CASIA, we obtain database $X$ with 12370 images, training set $X_{tr}$ with 2800 images, and testing set $X_s$ with 1400 images; for VGGFace2, we obtain database $X$ with 11413 images, training set $X_{tr}$ with 2800 images, and testing set $X_s$ with 1400 images. All images are resized to $224\times224$. Finally, we vote for different hash centers using $X_{tr}$.

\subsubsection{Models}For each dataset, we train the model based on 2 state-of-the-art deep hash algorithms (DHD \cite{jang2022deep} and CSQ \cite{yuan2020central}) using 2 different DNNs (VGG16 and ResNet50) as the feature layer and a fully connected layer with 64-dimensional output as the hash layer, to obtain 8 deep hash models with 64-bit hash code outputs. We also additionally train DHD-VGG19 and DHD-ResNet34 on CASIA for more testing in robustness experiments and ablation studies.

\subsubsection{Image post-processing operations}We use JPEG compression (JPEG), resizing (Resize), Gaussian blurring (Blur), rotation (Rotate), and Gaussian noise (Noise) to simulate the image post-processing of social platforms on the adversarial examples, and test their robustness, referring to \cite{zhang2024dual}. The default JPEG compression factor is $60$, the resize ratio is $3/2$, the Gaussian kernel size for Gaussian blur is $3$ with a standard deviation of $0.8$, the rotation angle is $10$ degrees, and the noise magnitude is $0.002^{0.5}$. We also manually upload the original images and adversarial examples to OSNs for post-processing in real-world scenarios and download them for testing, further evaluating the robustness of adversarial perturbations.

\subsubsection{Metrics}We use mean Average Precision (mAP) to measure retrieval precision, and a larger mAP means higher retrieval precision. Suppose the retrieval system returns the top-$k$ most similar images. If we perform $Q$ queries, with each query having $c_i (1\leq i\leq Q, 0\leq c_i\leq k)$ correct results, and the corresponding similarity rankings being $a_{i_1}, …, a_{i_{c_i}}$, the calculation of mAP is as follows:
\begin{equation}
mAP = \frac{1}{Q}{\sum_{i=1}^{Q} \frac{1}{c_i}}{\sum_{j=1}^{c_i} \frac{j}{a_{i_j}}}.\label{eq12}
\end{equation}

Similar to \cite{tang2024once}, we compute the mAP of the testing set on top 300. We record the mAP of the original images (O), the mAP of the adversarial examples (A), the mAP of the post-processed original images (PO), and the mAP of the post-processed adversarial examples (PA). A lower value of A or PA means better performance. We highlight the best experimental results in \textbf{bold} and labeled the second-best experimental results using \underline{underline}.

\subsubsection{Comparisons}We compare TOAP with existing state-of-the-art methods CWDM\cite{xiao2020evade}, DHTA\cite{bai2020targeted}, AdvHash \cite{hu2021advhash}, and UTAP \cite{tang2024once}. Since CWDM, DHTA, and AdvHash are not universal adversarial examples, we modify these methods appropriately for a fair comparison following \cite{tang2024once}. Specifically, since DHTA is a targeted attack aimed at the hash centers of other clusters, we modify its objective to target the negative hash centers of the original clusters. DHTA and CWDM generate specific perturbations for a single image, which leads us to follow the traditional universal adversarial perturbation generation method by summing up different perturbations for comparison. AdvHash is a targeted class-wise universal adversarial patch, similarly, we modify its objective to be the negative hash centers of the original clusters and sum up adversarial patch following its original settings.

\subsubsection{Implementation details}For TOAP, we set the number of iterations $T=100$, the hyperparameters $\alpha=0.3$, $ \beta=0.7$, and $\lambda_1:\lambda_2:\lambda_3 =1:10^{-3}:10^{-5}$. For VGGFace2-CSQ-VGG16 and VGGFace2-CSQ-ResNet50, the learning rate is 0.03; for the other models, the learning rate is 0.02. The pre-trained parameters of the U-Net module in CG are loaded from \cite{qu2024df}. For all methods, we set $\epsilon=16/255$ and set the patch percentage to be 0.05. We train the adversarial perturbations on $X_{tr}$ and evaluate them on $X_s$.

\subsection{Balancing Universality, Transferability, and Robustness}
We compare the universality, transferability, and robustness of the 5 different methods on different datasets, algorithms, models, and image post-processing operations, and record the mAPs of the adversarial examples before and after JPEG, Resize, Blur, Rotate, and Noise as shown in Table \ref{table2}. The data in the table are all obtained by testing a universal adversarial perturbation or patch applied to all images.

\begin{table*}[t]
\setlength{\tabcolsep}{3.7pt}
\renewcommand{\arraystretch}{1.18}
\caption{Quantitative comparison of universality, transferability, and robustness. All are mAPs (\%) of adversarial examples (A$\downarrow$, PA$\downarrow$) except \textbf{O/PO}, which is mAPs of the original images. * denotes the white-box adversarial results without post-processing.}
\label{table2}
\centering
\begin{tabular}{cccccccccc|ccccccc} 
\Xhline{1.1pt}
\multirow{2}{*}{\textbf{Dataset }} & \multirow{2}{*}{\textbf{Model} ~}& \multirow{2}{*}{\textbf{Method} ~} & \multicolumn{7}{c}{\textbf{VGG16 }} & \multicolumn{7}{c}{\textbf{ResNet50 }} \\ 
\cline{4-17}
 & & & \textbf{N/A} & \textbf{JPEG} & \textbf{Resize} & \textbf{Blur} & \textbf{Rotate} & \textbf{Noise} & \textbf{AVG} & \textbf{N/A} & \textbf{JPEG} & \textbf{Resize} & \textbf{Blur} & \textbf{Rotate} & \textbf{Noise} & \textbf{AVG} \\ 
\hline
\multicolumn{3}{c}{\textbf{Algorithm} ~ ~} & \multicolumn{14}{c}{\textbf{DHD}} \\ 
\hline
\multicolumn{1}{c}{\textbf{} ~ ~} &\multicolumn{1}{c}{\textbf{} ~ ~} &\multicolumn{1}{c}{\textbf{O/PO}} & 86.17 & 85.88 & 86.39 & 86.31 & 78.17 & 81.73 & 84.11 & 88.58 & 89.12 & 89.09 & 89.04 & 80.44 & 80.93 & 86.20 \\ 
\cline{2-17}
\multirow{10}{*}{CASIA }& \multirow{5}{*}{VGG16} & CWDM & 76.75* & 78.83 & 80.02 & 81.39 & 70.80 & 75.16 & 77.16 & 49.83 & 63.85 & 69.55 & 78.36 & 40.19 & 45.40 & 57.86 \\
 & & DHTA & 53.94* & 76.92 & 67.74 & 76.81 & 61.36 & 70.78 & 67.93 & 46.42 & 70.60 & 61.88 & 72.68 & 47.86 & 51.96 & 58.57 \\
 & & AdvHash & \uline{5.56*} & \uline{12.29} & 11.94 & 81.13 & 42.32 & \textbf{4.24} & 26.25 & 87.52 & 87.34 & 87.56 & 87.81 & 76.97 & 78.26 & 84.24 \\
 & & UTAP & 7.58* & 12.32 & \uline{10.01} & \uline{21.92} & \uline{10.02} & 10.15 & \uline{12.00} & \uline{21.67} & \uline{33.42} & \uline{27.19} & \uline{36.28} & \uline{27.72} & \uline{29.42} & \uline{29.28} \\ 
\cline{3-17}
 & & \textbf{TOAP}& \textbf{4.64*} & \textbf{9.57} & \textbf{6.62} & \textbf{14.30} & \textbf{7.63} & \uline{6.99} & \textbf{8.29} & \textbf{16.49} & \textbf{26.94} & \textbf{21.37} & \textbf{29.24} & \textbf{21.73} & \textbf{24.23} & \textbf{23.33} \\ 
\cline{2-17}
 & \multirow{5}{*}{ResNet50} & CWDM & 79.93 & 81.91 & 83.44 & 84.11 & 74.14 & 78.29 & 80.30 & 61.11* & 68.94 & 76.78 & 83.07 & 50.88 & 56.08 & 66.14 \\
 & & DHTA & 80.43 & 83.33 & 82.58 & 83.71 & 74.07 & 80.47 & 80.77 & 55.39* & 74.63 & 67.72 & 77.11 & 53.67 & 55.74 & 64.04 \\
 & & AdvHash & 82.20 & 82.36 & 81.96 & 83.84 & 74.01 & 75.25 & 79.94 & \textbf{3.63*} & \textbf{4.65} & \textbf{4.22} & 87.08 & 74.62 & \textbf{3.63} & 29.64 \\
 & & UTAP & \uline{35.12} & \uline{54.93} & \uline{44.26} & \uline{55.22} & \uline{44.65} & \uline{49.73} & \uline{47.32} & 11.73* & 22.78 & 15.66 & \uline{23.39} & \uline{15.33} & 15.16 & \uline{17.34} \\ 
\cline{3-17}
 & & \textbf{TOAP}& \textbf{19.46} & \textbf{32.39} & \textbf{25.17} & \textbf{33.27} & \textbf{25.88} & \textbf{27.39} & \textbf{27.26} & \uline{6.85*} & \uline{8.70} & \uline{7.40} & \textbf{8.30} & \textbf{8.36} & \uline{6.56} & \textbf{7.70} \\ 
\hline
\multicolumn{1}{c}{\textbf{} ~ ~} &\multicolumn{1}{c}{\textbf{} ~ ~} &\multicolumn{1}{c}{\textbf{O/PO}} & 93.37 & 93.39 & 93.27 & 93.42 & 89.12 & 89.61 & 92.03 & 94.79 & 94.84 & 94.97 & 95.01 & 89.62 & 86.15 & 92.56 \\ 
\cline{2-17}
\multirow{10}{*}{VGGFace2 }& \multirow{5}{*}{VGG16} & CWDM & 83.47* & 84.44 & 87.97 & 87.79 & 78.00 & 81.80 & 83.91 & 75.58 & 81.72 & 81.26 & 85.14 & 68.03 & 70.35 & 77.01 \\
 & & DHTA & 54.88* & 84.79 & 74.40 & 88.59 & 61.55 & 70.46 & 72.45 & 75.20 & 87.27 & 84.20 & 88.87 & 72.15 & 74.67 & 80.39 \\
 & & AdvHash & \textbf{6.09*} & \uline{19.47} & 22.53 & 88.78 & 50.20 & \textbf{6.71} & 32.30 & 92.61 & 92.19 & 93.31 & 94.17 & 80.34 & 79.00 & 88.60 \\
 & & UTAP & 10.83* & 25.10 & \uline{15.42} & \uline{44.87} & \uline{14.61} & 16.26 & \uline{21.18} & \uline{41.83} & \uline{63.83} & \uline{57.41} & \uline{67.55} & \uline{42.88} & \uline{54.10} & \uline{54.60} \\ 
\cline{3-17}
 & & \textbf{TOAP}& \uline{7.20*} & \textbf{8.91} & \textbf{7.78} & \textbf{11.52} & \textbf{8.27} & \uline{7.48} & \textbf{8.53} & \textbf{31.66} & \textbf{41.94} & \textbf{39.34} & \textbf{47.15} & \textbf{34.76} & \textbf{36.14} & \textbf{38.50} \\ 
\cline{2-17}
 & \multirow{5}{*}{ResNet50} & CWDM & 84.00 & 87.17 & 89.54 & 91.32 & 78.23 & 82.54 & 85.47 & 75.18* & 80.53 & 82.81 & 87.88 & 67.89 & 71.84 & 77.69 \\
 & & DHTA & 83.42 & 89.21 & 89.16 & 90.81 & 80.98 & 84.37 & 86.33 & 65.27* & 87.06 & 77.35 & 87.18 & 71.33 & 70.04 & 76.37 \\
 & & AdvHash & 89.39 & 90.52 & 90.79 & 91.81 & 84.35 & 84.20 & 88.51 & \textbf{3.57*} & \textbf{3.57} & \uline{9.79} & 90.77 & 76.32 & \textbf{3.57} & 31.27 \\
 & & UTAP & \uline{35.75} & \uline{54.38} & \uline{44.11} & \uline{54.88} & \uline{38.35} & \uline{41.14} & \uline{44.77} & 9.14* & 18.98 & 11.00 & \uline{18.68} & \uline{15.80} & 11.87 & \uline{14.25} \\ 
\cline{3-17}
 & & \textbf{TOAP}& \textbf{20.99} & \textbf{36.53} & \textbf{30.66} & \textbf{41.19} & \textbf{27.08} & \textbf{28.33} & \textbf{30.80} & \uline{7.08*} & \uline{8.00} & \textbf{7.08} & \textbf{8.10} & \textbf{7.99} & \uline{7.22} & \textbf{7.58} \\ 
\hline
\multicolumn{3}{c}{\textbf{Algorithm} ~ ~} & \multicolumn{14}{c}{\textbf{CSQ}} \\ 
\hline
\multicolumn{1}{c}{\textbf{} ~ ~} &\multicolumn{1}{c}{\textbf{} ~ ~} &\multicolumn{1}{c}{\textbf{O/PO}} & 83.42 & 83.66 & 83.12 & 82.55 & 76.99 & 72.70 & 80.41 & 87.54 & 87.19 & 87.32 & 86.96 & 79.42 & 79.25 & 84.61 \\ 
\cline{2-17}
\multirow{10}{*}{CASIA }& \multirow{5}{*}{VGG16} & CWDM & 54.37* & 61.87 & 64.08 & 73.17 & 45.38 & 51.57 & 58.41 & 70.15 & 72.21 & 74.75 & 76.84 & 63.07 & 65.96 & 70.50 \\
 & & DHTA & 34.27* & 62.55 & 54.93 & 74.45 & 39.70 & 44.94 & 51.81 & 60.08 & 74.40 & 69.75 & 77.28 & 65.77 & 68.50 & 69.30 \\
 & & AdvHash & \textbf{3.77*} & \textbf{5.56} & \uline{6.30} & 65.63 & 23.74 & \textbf{3.75} & 18.13 & 86.54 & 86.37 & 86.85 & 87.03 & 78.12 & 77.03 & 83.66 \\
 & & UTAP & 7.34* & \uline{7.88} & 8.12 & \textbf{17.65} & \uline{7.92} & 7.41 & \uline{9.39} & \uline{32.25} & \uline{48.61} & \uline{39.98} & \uline{55.19} & \uline{39.22} & \uline{43.50} & \uline{43.13} \\ 
\cline{3-17}
 & & \textbf{TOAP}& \uline{3.79*} & 8.93 & \textbf{5.86} & \uline{19.22} & \textbf{4.20} & \uline{5.17} & \textbf{7.86} & \textbf{22.47} & \textbf{29.20} & \textbf{27.46} & \textbf{34.98} & \textbf{19.63} & \textbf{29.82} & \textbf{27.26} \\ 
\cline{2-17}
 & \multirow{5}{*}{ResNet50} & CWDM & 58.89 & 64.69 & 68.87 & 76.31 & 51.22 & 53.60 & 62.26 & 69.71* & 74.83 & 76.81 & 79.62 & 68.64 & 67.34 & 72.83 \\
 & & DHTA & 60.36 & 72.97 & 73.32 & 78.43 & 55.41 & 59.86 & 66.73 & 42.64* & 77.40 & 61.02 & 74.64 & 66.42 & 61.06 & 63.86 \\
 & & AdvHash & 76.17 & 75.70 & 79.50 & 78.30 & 66.18 & 64.14 & 73.33 & \uline{3.65*} & 76.14 & 86.37 & 55.70 & 77.98 & \textbf{3.63} & 50.58 \\
 & & UTAP & \uline{25.36} & \uline{35.50} & \uline{35.19} & \uline{51.62} & \uline{31.15} & \uline{27.97} & \uline{34.47} & 3.74* & \uline{7.00} & \uline{4.58} & \uline{9.57} & \uline{6.32} & 4.64 & \uline{5.98} \\ 
\cline{3-17}
 & & \textbf{TOAP}& \textbf{17.83} & \textbf{29.29} & \textbf{26.40} & \textbf{43.21} & \textbf{18.63} & \textbf{21.27} & \textbf{26.11} & \textbf{3.61*} & \textbf{5.09} & \textbf{3.61} & \textbf{4.71} & \textbf{4.23} & \uline{3.85} & \textbf{4.18} \\ 
\hline
\multicolumn{1}{c}{\textbf{} ~ ~} &\multicolumn{1}{c}{\textbf{} ~ ~} &\multicolumn{1}{c}{\textbf{O/PO}} & 93.22 & 92.83 & 93.01 & 93.11 & 87.00 & 89.08 & 91.38 & 93.62 & 93.34 & 93.54 & 93.30 & 89.08 & 88.92 & 91.97 \\ 
\cline{2-17}
\multirow{10}{*}{VGGFace2 }& \multirow{5}{*}{VGG16} & CWDM & 78.19* & 82.51 & 83.32 & 86.57 & 71.41 & 77.68 & 79.95 & 83.35 & 85.41 & 86.42 & 90.00 & 75.52 & 79.80 & 83.42 \\
 & & DHTA & 32.36* & 72.09 & 56.37 & 85.27 & 35.95 & 48.23 & 55.05 & 75.63 & 86.14 & 83.01 & 88.09 & 77.86 & 79.82 & 81.76 \\
 & & AdvHash & \textbf{3.70*} & \textbf{5.40} & 6.11 & 84.26 & 34.66 & \textbf{3.69} & 22.97 & 92.91 & 92.40 & 92.71 & 92.32 & 86.48 & 86.72 & 90.59 \\
 & & UTAP & \uline{3.80*} & \uline{7.13} & \uline{5.16} & \textbf{14.87} & \uline{5.75} & 4.32 & \textbf{6.84} & \uline{35.10} & \textbf{54.08} & \uline{46.69} & \uline{60.91} & \textbf{39.12} & \uline{41.97} & \uline{46.31} \\ 
\cline{3-17}
 & & \textbf{TOAP}& \textbf{3.70*} & 7.48 & \textbf{4.78} & \uline{16.98} & \textbf{4.20} & \uline{4.21} & \uline{6.89} & \textbf{30.50} & \uline{60.05} & \textbf{43.31} & \textbf{60.90} & \uline{41.28} & \textbf{41.73} & \textbf{46.30} \\ 
\cline{2-17}
 & \multirow{5}{*}{ResNet50} & CWDM & 84.65 & 86.86 & 87.46 & 89.68 & 77.34 & 83.95 & 84.99 & 78.77* & 83.89 & 85.79 & 89.75 & 72.66 & 76.31 & 81.20 \\
 & & DHTA & 75.46 & 87.93 & 85.70 & 90.05 & 70.59 & 79.25 & 81.50 & 24.57* & 89.11 & 51.35 & 86.87 & 71.72 & 59.58 & 63.87 \\
 & & AdvHash & 87.15 & 87.90 & 88.78 & 89.48 & 79.33 & 82.47 & 85.85 & \textbf{3.59*} & 91.07 & 75.37 & 92.21 & 86.34 & \textbf{3.60} & 58.70 \\
 & & UTAP & \uline{52.85} & \uline{69.82} & \uline{65.34} & \uline{79.41} & \uline{56.25} & \uline{56.03} & \uline{63.28} & 4.13* & \uline{22.98} & \uline{8.01} & \uline{24.99} & \textbf{10.17} & \uline{7.41} & \uline{12.95} \\ 
\cline{3-17}
 & & \textbf{TOAP}& \textbf{25.30} & \textbf{42.56} & \textbf{36.17} & \textbf{51.16} & \textbf{26.37} & \textbf{26.19} & \textbf{34.63} & \uline{3.84*} & \textbf{13.68} & \textbf{5.26} & \textbf{12.87} & \uline{10.65} & 8.28 & \textbf{9.10} \\
\Xhline{1.1pt}
\end{tabular}

\end{table*}

\subsubsection{Universality}The white-box experimental results (*) for the adversarial examples before post-processing (N/A) demonstrate the universality of TOAP. It shows that TOAP comprehensively outperforms CWDM, DHTA as well as UTAP, and is comparable to AdvHash. However, AdvHash, a universal adversarial patch method, is less invisible and forces white-box models to focus more on the patch, thereby enhancing its adversariality.

\subsubsection{Transferability}The black-box experimental results for the adversarial examples before processing (N/A) demonstrate the transferability of TOAP. It shows that TOAP significantly outperforms existing methods and improves transferability by about 5\% to 28\%. AdvHash, although it has good white-box universality, its black-box transferability is not even as good as CWDM and DHTA with poor white-box universality. This could be attributed to the fact that the adversarial patches lack attack on global features, ignoring the differences in global features extracted by different models, which causes overfitting to the white-box model.

\subsubsection{Robustness}The experiments on post-processed adversarial examples (JPEG, Resize, Blur, Rotate, and Noise) demonstrate the robustness of TOAP. In the white-box models, we find that TOAP significantly outperforms the CWDM and DHTA, and is comparable to the state-of-the-art adversarial perturbation method UTAP and adversarial patch method AdvHash. In addition, we find that AdvHash has poor robustness under Blur and Rotate, while our method exhibits excellent robustness under all image post-processing operations. We also observe that AdvHash achieves better adversariality on VGG16 but performs poorly on ResNet50 across all settings of CSQ algorithm. This suggests that using ResNet50 as feature extractor provides better robustness against adversarial patch. In contrast, our TOAP demonstrates excellent adversariality on both VGG16 and ResNet50. In the best case (DHD-VGG16 on VGGFace2), the mAP of TOAP after Blur decreases by an additional 33\% compared to other methods, which demonstrates the strong robustness of our method.

In the black-box models, the experimental results can simultaneously demonstrate three properties of adversarial examples, i.e., universality, transferability, and robustness. Among 40 different cases, we have 38 cases that are significantly better than the existing second-best method and 2 cases that are slightly worse than UTAP, but still significantly better than CWDM, DHTA, and AdvHash. And it may be due to CG's insufficient simulation of various image post-processing operations in a few cases, leading to a slight performance decrease. In the best case (CSQ-ResNet50 to VGG16 on VGGFace2), the mAP of TOAP after Rotate decreases by about 30\% compared to the second-best methods, further validating the robustness of our TOAP.

Finally, we calculate the average value of all mAPs (AVG). The experimental results demonstrate that our method achieves the best balance of universality, transferability, and robustness in 15 cases, and is slightly worse than UTAP in 1 case, but still significantly better than CWDM, DHTA, and AdvHash. In the black-box experiments with CSQ-ResNet50 on VGGFace2, the average adversariality of TOAP surpasses the second-best method by about 29\%, demonstrating that our method effectively balances universality, transferability, and robustness of adversarial perturbation.

\subsection{Additional Robustness Evaluation}
To conduct a more comprehensive robustness analysis, we select the CASIA dataset and generate adversarial examples under DHD-VGG16 for further testing and analysis.

\begin{figure*}[tb]
 \centering
 \includegraphics[width=1.0\linewidth]{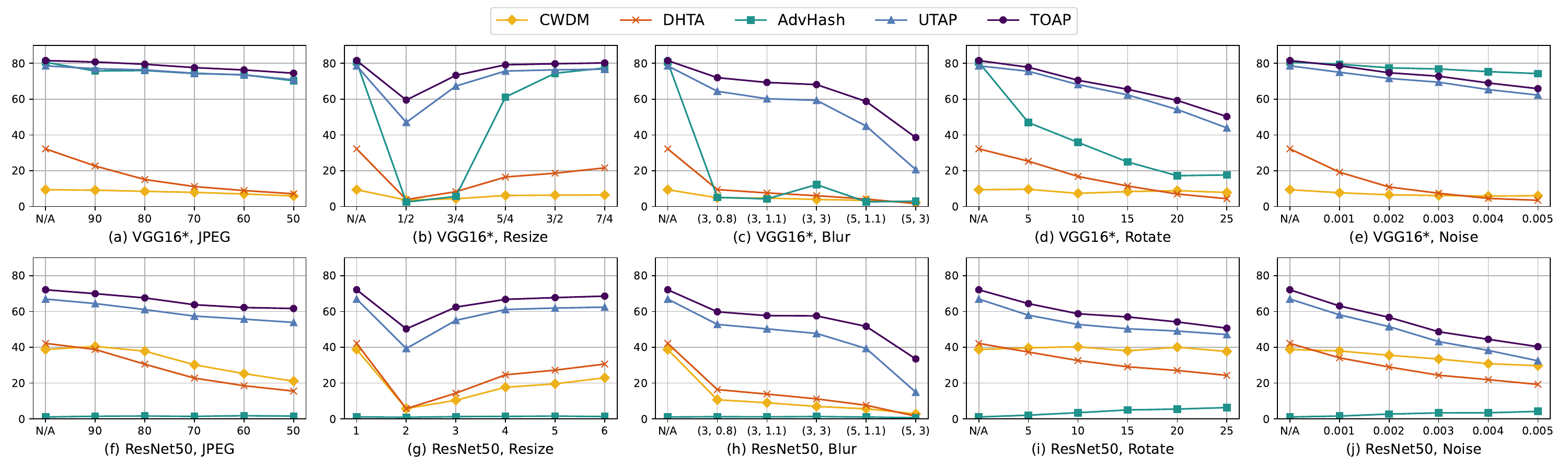}
 \caption{Robustness evaluation of adversarial examples after different image post-processing operations, Y-axis denotes $\Delta$mAP$\uparrow$. For JPEG, X-axis denotes different compression factors. For Resize, X-axis denotes different resize ratios. For Blur, X-axis denotes different Gaussian kernels and standard deviation. For Rotate, X-axis represents different rotation angles. For Noise, X-axis denotes different noise magnitudes, and for the sake of concise representation, we abbreviate $x^{0.5}$ to x. The experimental setting is CASIA, DHD, VGG16.}
 \label{fig9}
\end{figure*}

\subsubsection{Robustness under Extensive Post-processing Operations}For the purpose of comprehensively evaluating the robustness of TOAP, we further analyze the robustness of adversarial examples against various levels of JPEG, Resize, Blur, Rotate, and Noise. We test the adversarial examples on DHD-VGG16 and DHD-ResNet50, respectively, demonstrating the robustness of our universal perturbation in white-box and black-box models. For ease of comparison, we calculate the change in mAP before and after adding perturbation or patch (denoted as $\Delta$mAP) to represent the adversariality of different methods, the results are shown in Fig. \ref{fig9}. Y-axis represents $\Delta$mAP and X-axis represents different post-processing parameters. A larger $\Delta$mAP indicates better performance. Fig. \ref{fig9}(a)$\sim$(e) show the robustness experimental results in the white-box model and Fig. \ref{fig9}(f)$\sim$(j) show those in the black-box model.

It is discovered that, in the white-box, with JPEG compression, Resize, Gaussian Blur, and Rotate on the image, the $\Delta$mAP of TOAP outperforms all other methods, demonstrating the superior robustness of TOAP. When Gaussian noise is added, AdvHash performs slightly better than TOAP. This may be due to the addition of whole-image random perturbations, which enhances the performance of adversarial patch. This also confirms our hypothesis that adversarial perturbation on global features has better performance compared to local adversarial patch. For the black-box, TOAP significantly outperforms all comparison methods under different image post-processing operations. The AdvHash method, which performs well in the white-box after JPEG and Noise, is also not robust in the black-box due to its poor transferability.

Notably, the performance of CWDM and DHTA in the black-box is slightly better than in the white-box, both for the adversarial examples before and after post-processing. This may be caused by the differences in feature processing between VGG16 and ResNet50. VGG16 is more biased towards learning localized and low-level features, whereas the residual blocks of ResNet50 allow it to learn more global and high-level features. This means that while adversarial examples perform poorly in VGG16, these adversarial perturbations may be recognized in ResNet50 successfully due to the more complex and high-level feature representation of ResNet50. Meanwhile, a series of image post-processing usually introduces noise or leads to information loss, affecting the features of the image. VGG16, due to its low-level and simple feature extraction method, may be more susceptible to these noises and information loss, causing it to fail to recognize the perturbations, whereas the deeper structure and residual connections of ResNet50 allow it to better extract the adversarial information, resulting in higher adversariality. In contrast, TOAP can be successfully recognized by VGG16 and ResNet50, which shows stronger robustness.

\subsubsection{Robustness under OSNs}
\begin{table}
\centering
\setlength{\tabcolsep}{4pt}
\renewcommand{\arraystretch}{1.25}
\setlength{\abovecaptionskip}{3.3pt} 
\setlength{\belowcaptionskip}{3.3pt} 
\caption{Robustness Evaluation (mAP\%) under OSNs. The experimental setting is CASIA, DHD, VGG16. All are mAPs of adversarial examples (PA$\downarrow$) except \textbf{PO}.}
\label{table3}
\begin{tabular}{ccccccc} 
\Xhline{1.1pt}
\textbf{OSNs}& \textbf{Method}& \textbf{VGG16}& \textbf{VGG19}& \textbf{ResNet34}& \textbf{ResNet50}& \textbf{AVG}\\ 
\hline
\multicolumn{1}{c}{\textbf{} ~ ~} &\multicolumn{1}{c}{\textbf{PO}} & 82.95 & 84.36 & 86.67 & 83.20 & 84.30 \\ 
\cline{2-7}
\multirow{4}{*}{Facebook}& CWDM & 78.80 & 77.14 & 79.02 & 74.89 & 77.46 \\
 & DHTA & 80.91 & 76.07 & 79.27 & 79.09 & 78.84 \\
 & AdvHash & \uline{12.26} & 74.65 & 84.79 & 84.51 & 64.05 \\
 & UTAP & 16.40 & \uline{28.01} & \uline{47.72} & \uline{39.75} & \uline{32.97} \\ 
\cline{2-7}
 & \textbf{TOAP}& \textbf{9.88} & \textbf{20.50} & \textbf{35.50} & \textbf{26.74} & \textbf{23.16} \\ 
\hline
\multicolumn{1}{c}{\textbf{} ~ ~} &\multicolumn{1}{c}{\textbf{PO}} & 79.78 & 84.13 & 84.51 & 82.42 & 82.71 \\ 
\cline{2-7}
\multirow{4}{*}{WeChat}& CWDM & 74.92 & 76.25 & 75.89 & 60.21 & 71.82 \\
 & DHTA & 76.58 & 77.95 & 81.70 & 75.34 & 77.89 \\
 & AdvHash & \uline{19.25} & 79.67 & 83.21 & 81.10 & 65.81 \\
 & UTAP & 22.08 & \uline{30.90} & \uline{45.60} & \uline{36.33} & \uline{33.73} \\ 
\cline{2-7}
 & \textbf{TOAP}& \textbf{13.96} & \textbf{23.78} & \textbf{40.09} & \textbf{26.61} & \textbf{26.11} \\ 
\hline
\multicolumn{1}{c}{\textbf{} ~ ~} &\multicolumn{1}{c}{\textbf{PO}} & 83.28 & 84.03 & 86.28 & 85.43 & 84.76 \\ 
\cline{2-7}
\multirow{4}{*}{Weibo}& CWDM & 75.74 & 75.95 & 72.32 & 51.32 & 68.83 \\
 & DHTA & 69.78 & 71.56 & 71.86 & 64.42 & 69.41 \\
 & AdvHash & 7.77 & 73.87 & 85.29 & 83.84 & 62.69 \\
 & UTAP & \uline{6.70} & \uline{19.32} & \textbf{23.60} & \uline{25.52} & \uline{18.79} \\ 
\cline{2-7}
 & \textbf{TOAP}& \textbf{5.30} & \textbf{17.47} & \uline{25.67} & \textbf{21.71} & \textbf{17.54} \\
\Xhline{1.1pt}
\end{tabular}
\end{table}
To further evaluate the robustness of TOAP, we use OSNs to process the images realistically, including Facebook, WeChat, and Weibo, which are the most commonly used nowadays. We select the CASIA dataset and generate adversarial examples in DHD-VGG16. For different platforms and methods, we manually upload and download 84 universal adversarial examples for evaluating respectively. Then, we test the performance of these downloaded adversarial examples on DHD-VGG16, -VGG19, -ResNet34, and -ResNet50. Table \ref{table3} shows the results.

The experimental results demonstrate that TOAP outperforms the second-best method in 11 out of 12 cases. And in 1 case, TOAP performs slightly worse than UTAP, but remains clearly superior overall. In the best case (Facebook, DHD, ResNet50), TOAP outperforms the second-best method by about 13\%. Notably, our method maintains good robustness even under the highest compression of WeChat.

Finally, we calculate the average value of mAPs under different models to analyze the advantage of TOAP more clearly. It can be found that the average adversarial mAP of our method under Facebook, WeChat, and Weibo is superior to other methods. In the best case, TOAP performs about 10\% better than the second-best method, which further validates the robustness of our method in reality.

\subsection{Visual Analysis}

\begin{figure}[tb]
 \centering
 \includegraphics[width=0.98\linewidth]{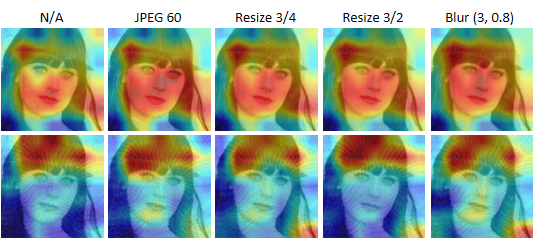}
 \caption{Visualization of the change in model’s focus areas of original images (line 1) and TOAP (line 2) before and after different image post-processing operations. The experimental setting is CASIA, DHD, VGG16.}
 \label{fig10}
\end{figure}

Firstly, we visualize the model's focus areas of the images before and after different image post-processing operations, as shown in Fig. \ref{fig10}. The line 1 shows the model's focus areas of the original image, which is mainly concentrated on the facial region. The line 2 shows the model's focus areas of TOAP, where the model almost entirely shifts its focus away from the facial region, causing the deep hash model fail to retrieve. We apply various image post-processing operations to both the original and adversarial images, including JPEG, Resize, and Blur. It can be observed that, after applying these different post-processing operations, the model's focus areas of both the original image and TOAP remain nearly unchanged, further validating the robustness of TOAP.

\begin{figure}[tb]
 \centering
 \includegraphics[width=1.0\linewidth]{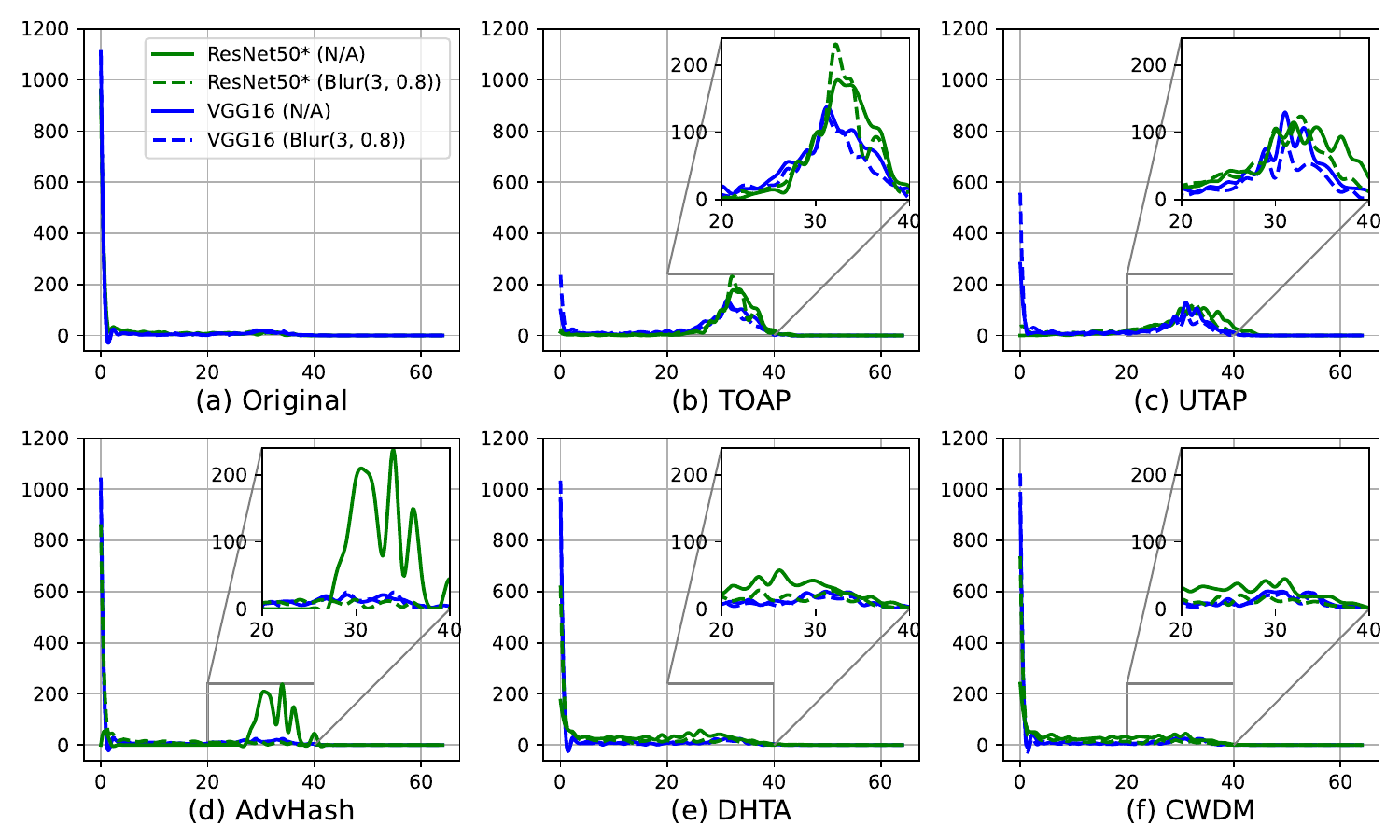}
 \caption{The relationship between the number of adversarial examples and the Hamming distance to their original cluster centers before and after Blur. A larger number of adversarial examples deviating far from 0 indicates stronger adversariality. The experimental setting is CASIA, DHD, ResNet50. * denotes the white-box adversarial results. For Blur, the Gaussian kernel size is 3 and the standard deviation is 0.8.}
 \label{fig11}
\end{figure}

Then, we visualize the relationship between the number of adversarial examples generated by different methods and the Hamming distance to their original cluster centers in different deep hash models before and after post-processing, as shown in Fig. \ref{fig11}. We magnify the gathering region of adversarial examples, where the Hamming distance is in $[20, 40]$. It can be observed that, in the white-box model and before post-processing (the green solid lines), both TOAP and AdvHash manage to move far away from their original clusters and concentrate in regions with a Hamming distance of [20, 40] from the original cluster. UTAP has fewer adversarial examples moving away from the original cluster, while DHTA and CWDM can hardly move away from the original cluster. After post-processing, AdvHash struggles to maintain its adversariality with more adversarial examples gathering in the original cluster, while TOAP maintains its adversariality well, as shown by the green dashed lines. In the black-box model (the blue solid and dashed lines), TOAP still manages to stay far from the original cluster with significantly fewer invalid adversarial examples gathered near the original cluster, compared to other methods. This further demonstrates that TOAP possesses the strongest transferability and robustness.

\begin{figure*}[tb]
	\centering
	\includegraphics[width=0.9\textwidth]{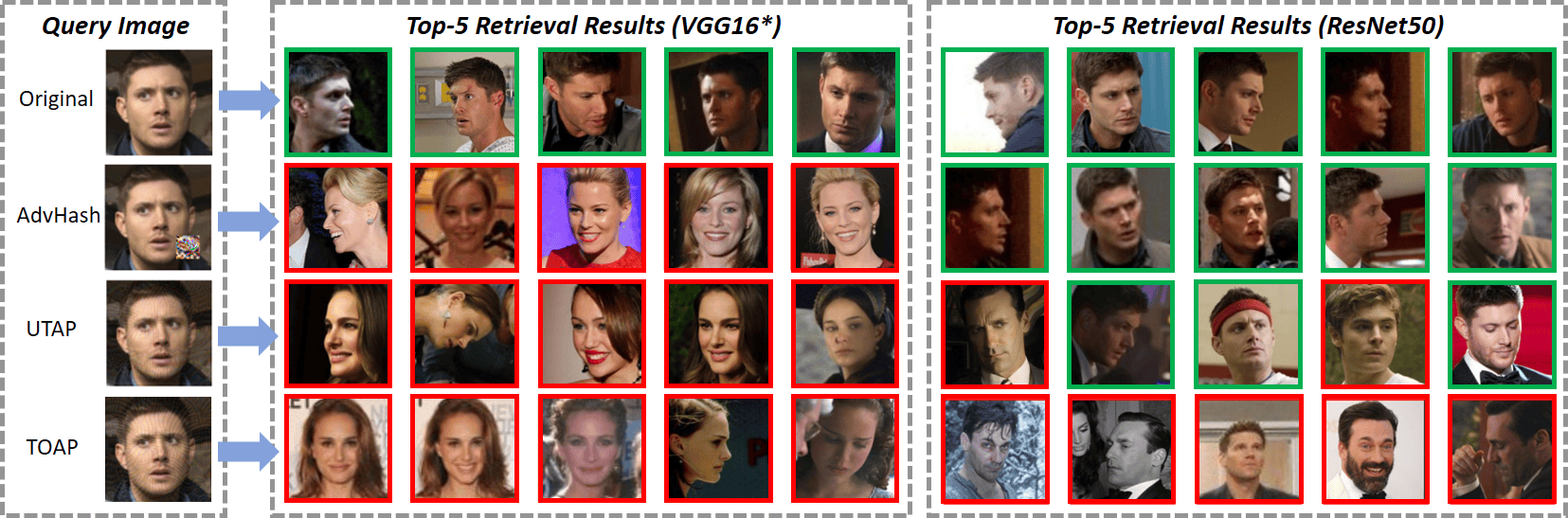} %
	\caption{Top-5 retrieval results of AdvHash, UTAP, and TOAP. The query images are post-processed by Facebook. The experimental setting is CASIA, DHD. * denotes the white-box model. The green boxes indicate the retrieval of the original identity, while red boxes indicate the retrieval of irrelevant identities.}
	\label{fig12}
\end{figure*}

Finally, we test the adversarial examples post-processed by Facebook. We compare TOAP with AdvHash and UTAP, visualize the top-5 retrieval results as shown in Fig. \ref{fig12}. It can be found that AdvHash, UTAP, and TOAP all exhibit excellent white-box performance. However, for the black-box model, AdvHash has no effect, and UTAP has better performance, but is still clearly inferior to TOAP.

\subsection{Ablation Studies}
We use the CASIA dataset to perform ablation experiments under different models and algorithms, enriching the experimental setup and proving the effectiveness of TOAP.

\begin{table}
\setlength{\tabcolsep}{5.8pt} 
\renewcommand{\arraystretch}{1.25} 
\setlength{\abovecaptionskip}{3.3pt} 
\setlength{\belowcaptionskip}{3.3pt} 
\caption{Ablation study for local and global CG (mAP\%). The dataset is CASIA. All are mAPs of adversarial examples (A$\downarrow$, PA$\downarrow$) except \textbf{O/PO}.}
\label{table4}
\centering
\begin{tabular}{cccccc} 
\Xhline{1.1pt}
\textbf{Alg-Model}& \textbf{Method}& \multicolumn{2}{c}{\textbf{Alg-VGG16}} & \multicolumn{2}{c}{\textbf{Alg-ResNet50}} \\ 
\hline
\multicolumn{1}{c}{\textbf{} ~ ~} &\multicolumn{1}{c}{\textbf{O/PO}} & O:86.17 & PO:83.70& O:88.58 & PO:85.72\\ 
\cline{2-6}
\multirow{2}{*}{\begin{tabular}[c]{@{}c@{}}DHD-\\ ResNet50\end{tabular}} & Global & \uline{20.56} & \uline{33.65} & \uline{8.10*}& 11.05 \\
& Local & 24.27 & 34.13 & 8.71*& \uline{9.61} \\ 
\cline{2-6}
& \textbf{TOAP}& \textbf{19.46} & \textbf{28.82} & \textbf{6.85*}& \textbf{7.86} \\ 
\hline
\multicolumn{1}{c}{\textbf{} ~ ~} &\multicolumn{1}{c}{\textbf{O/PO}} & O:83.42 & PO:79.80& O:87.54 & PO:84.03\\ 
\cline{2-6}
\multirow{2}{*}{\begin{tabular}[c]{@{}c@{}}CSQ-\\ VGG16\end{tabular}} & Global & \uline{4.71*}& 14.14 & \uline{29.00} & \uline{36.82} \\
& Local & 5.04*& \uline{13.64} & 29.09 & 40.86 \\ 
\cline{2-6}
& \textbf{TOAP}& \textbf{3.79*}& \textbf{8.68} & \textbf{22.47} & \textbf{28.22} \\
\Xhline{1.1pt}
\end{tabular}
\end{table}

Firstly, to demonstrate the effectiveness of local and global CG, we design three comparison methods: training adversarial perturbation using global CG, local CG as well as local and global CG, i.e., TOAP, and the results are shown in Table \ref{table4}. We conduct experiments on DHD-ResNet50 and CSQ-VGG16, and record the original mAP (O) and the adversarial mAP (A), as well as the average original/adversarial mAP after applying 5 default post-processing operations (PO/PA). The experimental results show that using local or global CG only may not process the image effectively enough, resulting in suboptimal performance of adversarial examples. And full consideration of local and global features can better improve the performance of adversarial examples.

Secondly, we verify the effectiveness of our optimization objectives. We generate adversarial perturbation based on different objectives on CSQ-ResNet50, and the results are shown in Table \ref{table5}. LS denotes away from the original cluster centers, LC denotes away from the data space centers, LO denotes away from the overall center, and other denotes the combination of two optimization objectives using meta-learning approach. The results show that optimizing perturbation using a single overall center for all images leads to optimization difficulty (LO), and while meta-learning can alleviate this, it can lead to higher performance if other optimization objectives are used directly instead of the overall center, i.e., TOAP. In addition, although LC performs well in the white-box model ResNet50 and ResNet34, which shares a similar structure to ResNet50, its transferability is limited in the more challenging VGG model with an unknown structure. In contrast, TOAP exhibits better transferability in VGG.

\begin{table}
\centering
\setlength{\tabcolsep}{6pt}
\renewcommand{\arraystretch}{1.25}
\setlength{\abovecaptionskip}{3.3pt} 
\setlength{\belowcaptionskip}{3.3pt} 
\caption{Ablation study for optimization objective (mAP\%). The experimental setting is CASIA, CSQ, ResNet50. All are mAPs of adversarial examples (A$\downarrow$) except \textbf{O}.}
\label{table5}
\begin{tabular}{cccccc} 
\Xhline{1.1pt}
\multirow{1}{*}{\textbf{Method}}& \textbf{ResNet34}& \textbf{ResNet50}& \textbf{VGG16}& \textbf{VGG19}& \textbf{AVG}\\ 
\cline{1-6}
\multirow{1}{*}{\textbf{O}} & 88.39 & 87.54 & 83.42 & 82.08 & 85.36 \\ 
\hline
LS& 20.95 & \textbf{3.73*}& \uline{25.77} & 18.62 & 17.27 \\
LC& \textbf{11.66} & \textbf{3.73*}& 26.46 & \uline{15.22} & \uline{14.27} \\
LO& 31.54 & 10.84*& 35.13 & 33.22 & 27.68 \\
LC+LO& \uline{13.59} & 3.88*& 26.53 & 15.25 & 14.81 \\
LS+LO& 29.15 & 3.81*& 32.79 & 30.10 & 23.96 \\ 
\hline
\textbf{LS+LC}& 16.18 & \uline{3.80*}& \textbf{21.28} & \textbf{13.84} & \textbf{13.78} \\
\Xhline{1.1pt}
\end{tabular}
\end{table}

\begin{figure}[tb]
 \centering
 \includegraphics[width=1\linewidth]{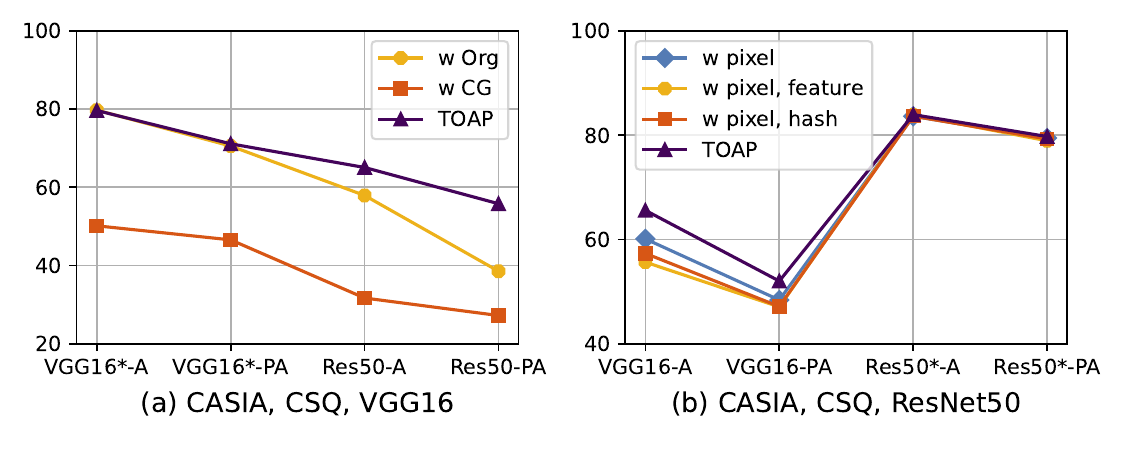}
 \caption{Ablation study for (a) $\mathcal{L}_{AP}$, (b) $\mathcal{L}_{CG}$. Y-axis represents $\Delta $mAP$\uparrow$ for the adversarial examples before (-A) and after (-PA) post-processing. The white-box model is emphasized with *.}
 \label{fig13}
\end{figure}

Thirdly, we perform ablation experiments on $\mathcal L_{AP}$ and the results are shown in Fig. \ref{fig13}(a). The adversarial examples are generated using original perturbed images (w Org), local and global CG-processed perturbed images (w CG), as well as both types of images (TOAP) on CSQ-VGG16. The experimental results show that, in the black-box model, TOAP significantly outperforms the other methods, which demonstrates that using both simultaneously for training can achieve the optimal balance of universality, transferability, and robustness.

Finally, we perform ablation experiments on the fine-tuning loss 
$\mathcal L_{CG}$. We generate adversarial examples on CSQ-ResNet50 and compare the $\Delta$mAP, as shown in Fig. \ref{fig13}(b). The experimental results show that TOAP has a significant advantage in the more challenging black-box model, which outperforms the other methods. This may be due to the fact that TOAP, fine-tuned by joint pixel-, feature-, and hash-level losses, is able to optimize both the low-level pixel features and high-level semantic features of images, enabling the simulation of OSNs with stronger generalization ability.

\subsection{Applications}

\begin{figure}[tb]
 \centering
 \includegraphics[width=0.98\linewidth]{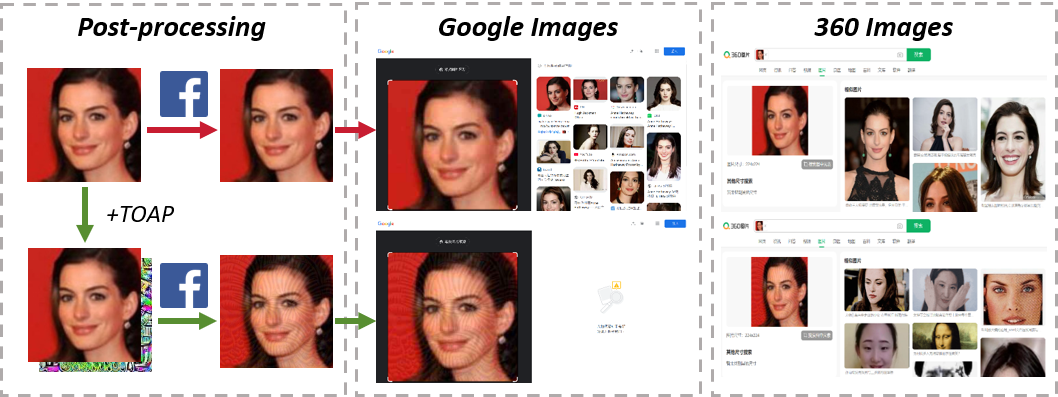}
 \caption{Illustration of TOAP protecting users' facial images from malicious retrieval in commercial applications.}
 \label{fig14}
\end{figure}

We further apply the proposed TOAP to real-world retrieval platforms, including Google Images\footnote{\url{https://images.google.com/}} and 360 Images\footnote{\url{https://st.so.com/}}. As shown in Fig. \ref{fig14}, we upload both the unperturbed and perturbed user images to Facebook for post-processing, and then use the post-processed images to conduct retrieval tests on Google Images and 360 Images. The results show that image with TOAP added fails to return retrieval results in Google Images and is unable to retrieve relevant user photos in 360 Images, demonstrating the effectiveness of our TOAP, even in real-world applications.

\section{CONCLUSIONS}

We propose TOAP, the first OSNs-oriented adversarial perturbation with simultaneous cross-image universality, cross-model transferability, and cross-image post-processing robustness, protecting users’ facial images from malicious retrieval. We propose local and global CG, a image post-processing method to simulate OSNs. Then, we define the optimization objectives as cluster centers and data space centers to optimize TOAP, which effectively addresses the optimization difficulty caused by a single objective. By alternately optimizing the adversarial examples using original as well as local and global CG-processed perturbed images, and training the CG using pixel-, feature-, and hash-level losses, we generate robust perturbation successfully. Extensive experiments demonstrate that TOAP is more robust in universal transferable anti-facial retrieval, which is able to successfully resist the image post-processing in OSNs and is more valuable for practical applications.




\normalem
\bibliographystyle{IEEEtran}
\bibliography{ref}

\end{document}